\documentclass[manuscript,screen]{acmart}

\AtBeginDocument{%
  \providecommand\BibTeX{{%
    \normalfont B\kern-0.5em{\scshape i\kern-0.25em b}\kern-0.8em\TeX}}}

\setcopyright{acmcopyright}
\copyrightyear{2023}
\acmYear{2023}
\acmDOI{XXXXXXX.XXXXXXX}

\usepackage{adjustbox}
\usepackage{csquotes}
\usepackage{multirow}
\usepackage{pifont}
\newcommand{\cmark}{\ding{51}}
\newcommand{\xmark}{\ding{55}}
\usepackage{makecell}
\usepackage{caption}
\usepackage{subcaption}

\begin{document}

\title{Redefining Counterfactual Explanations for Reinforcement Learning: Overview, Challenges and Opportunities}

\author{Jasmina Gajcin}
\email{gajcinj@tcd.ie}
\orcid{0000-0002-8731-1236}
\author{Ivana Dusparic}
\email{ivana.dusparic@tcd.ie}
\orcid{0000-0003-0621-5400}
\affiliation{%
  \institution{Trinity College Dublin}
  \streetaddress{College Green}
  \city{Dublin}
  \state{}
  \country{Ireland}
  \postcode{D02}
}

\renewcommand{\shortauthors}{Gajcin and Dusparic}

\begin{abstract}
  
While AI algorithms have shown remarkable success in various fields, their lack of transparency hinders their application to real-life tasks. Although explanations targeted at non-experts are necessary for user trust and human-AI collaboration, the majority of explanation methods for AI are focused on developers and expert users. Counterfactual explanations are local explanations that offer users advice on what can be changed in the input for the output of the black-box model to change. Counterfactuals are user-friendly and provide actionable advice for achieving the desired output from the AI system. While extensively researched in supervised learning, there are few methods applying them to reinforcement learning (RL). In this work, we explore the reasons for the underrepresentation of a powerful explanation method in RL. We start by reviewing the current work in counterfactual explanations in supervised learning. Additionally, we explore the differences between counterfactual explanations in supervised learning and RL and identify the main challenges that prevent the adoption of methods from supervised in reinforcement learning. Finally, we redefine counterfactuals for RL and propose research directions for implementing counterfactuals in RL.
\end{abstract}

\ccsdesc[500]{Computing methodologies~Reinforcement learning}

\keywords{Reinforcement Learning, Explainability, Interpretability, Counterfactual Explanations}

\maketitle

\section{Introduction}

Artificial intelligence (AI) solutions have become pervasive in various fields in the last decades, thanks in part to the adoption of deep learning algorithms. In particular, deep learning has shown remarkable success in supervised learning tasks, where the goal is to learn patterns in a labeled training data set and use them to accurately predict labels on unseen data \cite{pouyanfar2018survey}. Deep learning algorithms rely on neural networks, which allow for efficient processing of large amounts of unstructured data. However, they also rely on a large number of parameters, making their decision-making process difficult to understand. These models are often referred to as black-box due to the lack of transparency in their inner workings.

Reinforcement learning (RL) \cite{sutton2018reinforcement} is a sub-field of AI that focuses on developing intelligent agents for sequential decision-making tasks. RL employs a trial-and-error learning approach in which an agent learns a task from scratch through interactions with its environment. An agent can observe the environment, perform actions that alter its state and receive rewards from the environment, which guide it towards an optimal behavior. The goal of RL is to obtain an optimal policy $\pi$, which maps the agent's states to optimal actions. This bears some similarity to supervised learning approaches, where the goal is to classify an instance into the correct class according to the input features. However, while supervised learning algorithms rely on labeled training instances to learn patterns in the data, RL agents approach the task without prior knowledge and learn it through interactions with the environment. Deep reinforcement learning (DRL) algorithms \cite{8103164}, which employ a neural network to represent an agent's policy, are currently the most popular approach for learning RL policies \cite{8103164}. DRL algorithms have shown remarkable success in navigating sequential decision-making problems in games, autonomous driving, healthcare, and robotics \cite{mnih2013playing, kiran2021deep, coronato2020reinforcement, kober2013reinforcement}. Although they can process large amounts of unstructured, high-dimensional data, they also struggle to explain agent's decisions, due to their reliance on neural networks.

\begin{figure}[t]
    \centering
    \includegraphics[width=0.7\textwidth]{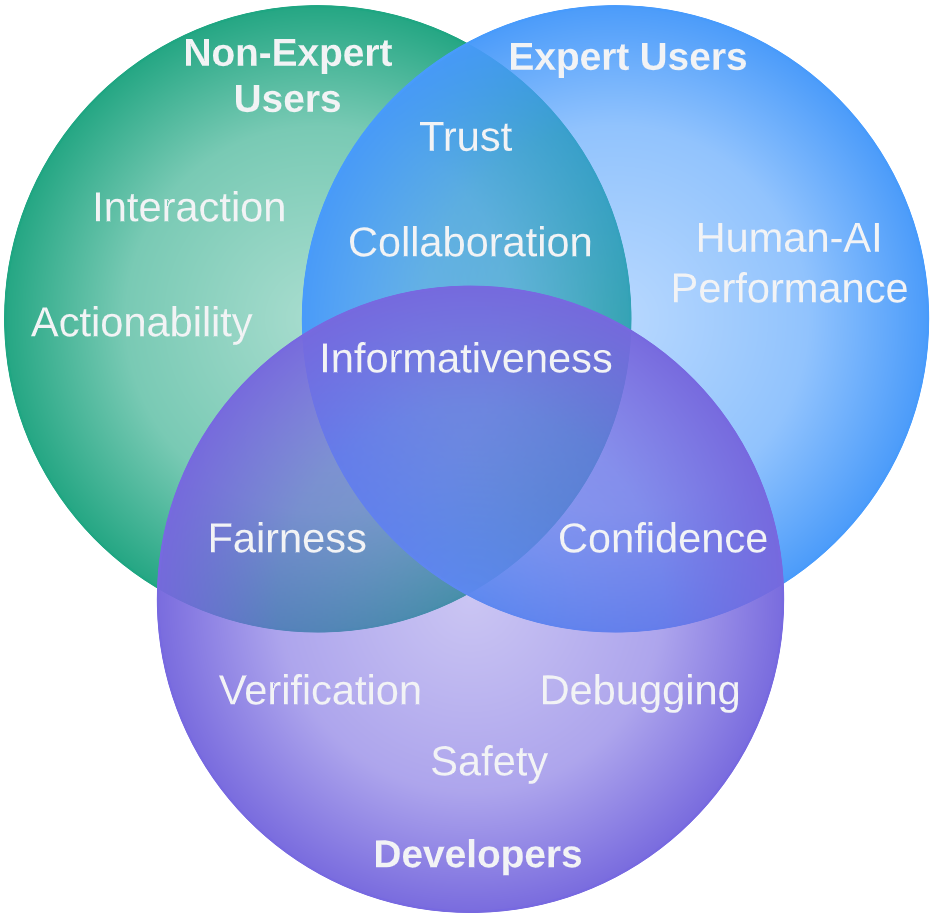}
    \caption{{A summary of goals of XAI depending on the target audience: while the focus of developers is to better understand the abilities of the system and enable successful deployment, experts using the system require explanations to better collaborate with the system. Explanations are necessary for non-expert users to develop trust, ensure system decisions are fair, and give users actionable feedback on how to elicit a different decision from the system.}}
    \label{taxonomy}
\end{figure}

Depending on the task and the user, AI systems require explainability for a variety of reasons (Figure 1). From the perspective of the developer, explainability is necessary to verify the system's behavior before deployment. Understanding how the input features influence the decision of the AI system is necessary to avoid deployment of models that rely on spurious correlations \cite{beery2018recognition,gajcin2022reccover} and to ensure robustness to adversarial attacks \cite{szegedy2013intriguing}. From the perspective of fairness, understanding the decision-making process of an AI system is necessary to prevent automated discrimination. Lack of transparency of AI models can cause them to inadvertently adopt historical biases ingrained in training data and use them in their decision logic \cite{jentzsch2019semantics}. To prevent discrimination, users of autonomous decision-making systems are now legally entitled to an explanation under the regulation within the GDPR in EU \cite{goodman2017european}. From the perspective of expert and non-expert users of the system, explainability is necessary to ensure trust. For experts that use AI systems as an aid in their everyday tasks, trust is a crucial component necessary for successful collaboration. For example, a medical doctor using an AI system for diagnostics needs to understand it to trust its decisions and use them for this high-risk task \cite{kundu2021ai}. Similarly, for non-expert users, trust is needed to encourage interaction with the system. If an AI system is used to make potentially life-altering decisions for the user, they need to understand how the system operates to maintain their confidence and trust in the system.

The field of explainable AI (XAI) explores methods for interpreting decisions of black-box systems in various fields such as machine learning, reinforcement learning, explainable planning \cite{lundberg2017unified, vstrumbelj2014explaining,ribeiro2016should, lakkaraju2017interpretable, frosst2017distilling,greydanus2018visualizing, puri2019explain, wang2016dueling, verma2019verifiable, verma2018programmatically,coppens2019distilling, amir2018highlights,sequeira2020interestingness}. In recent years, the focus of XAI has mostly been on explaining decisions of supervised learning models \cite{burkart2021survey}. Specifically, the majority of XAI methods have focused on explaining the decisions of neural networks, due to the emergence of deep learning as the state-of-the-art approach to many supervised learning tasks \cite{cao2018deep,abiodun2018state,yue2018machine}. In contrast, explainable RL (XRL) is a fairly novel field that has not yet received an equal amount of attention. Most often, existing XRL methods focus on explaining DRL algorithms, which rely on neural networks to represent the agent's policy, due to their prevalence and success \cite{vouros2022explainable}. However, as RL algorithms are becoming more prominent and are being considered for use in real-life tasks, there is a need for understanding their decisions \cite{puiutta2020explainable, dulac2019challenges}. For example, RL algorithms are being developed for different tasks in healthcare, such as dynamic treatment design \cite{zhao2009reinforcement,luckett2019estimating,li2019reinforcement}. Without rigorous verification and understanding of such systems, the medical experts will be reluctant to collaborate and rely on them \cite{riachi2021challenges}. Similarly, RL algorithms have been explored for enabling autonomous driving \cite{aradi2020survey}. To understand and prevent mistakes such as the 2017 Uber accident \cite{knight_2018} where a self-driving car failed to stop before a pedestrian, the underlying decision-making systems have to be scrutable. Specific to the RL framework, explainability is also necessary to correct and prevent ``reward hacking'' -- a phenomenon where an RL agent learns to trick a potentially misspecified reward function, such as a vacuum cleaner ejecting collected dust to increase its cleaning time \cite{amodei2016concrete,pan2022effects}.

In this work, we explore counterfactual explanations in supervised and reinforcement learning. Counterfactual explanations answer the question: \textit{``Given that the black-box model made decision y for input x, how can x be changed for the model to output alternative decision y'?''} \cite{verma2021counterfactual}.
Counterfactual explanations offer actionable advice to users of black-box systems by generating \textit{counterfactual instances} -- instances as similar as possible to the original instance being explained but producing a desired outcome. If the user is not satisfied with the decision of a black-box system, a counterfactual explanation offers them a recipe for altering their input features, to obtain a different output. For example, if a user is denied a loan by an AI system, they might be interested to know how they can change their application so that it gets accepted in the future. Counterfactual explanations are targeted at non-expert users, as they often deal in high-level terms, and offer actionable advice to the user. They are also selective, aiming to change as few features as possible to achieve the desired output. As explanations that can suggest potentially life-altering actions to the users, counterfactuals carry great responsibility. A useful counterfactual explanation can help user achieve a desired outcome, and increase their trust and confidence in the system. However, an ill-defined counterfactual that proposes unrealistic changes to the input features or does not deliver the desired outcome can waste user's time and effort and erode their trust in the system. For this reason, careful selection of counterfactual explanations is essential for maintaining user trust and encouraging their collaboration with the system. 

Although they have been explored in supervised learning \cite{dandl2020multi, chen2021relace, joshi2019towards, looveren2021interpretable,poyiadzi2020face, wachter2017counterfactual}, counterfactual explanations are rarely applied to RL tasks \cite{olson2019counterfactual}. In supervised learning, methods for generating counterfactual explanations often follow a similar pattern. Firstly, a loss function is defined, taking into account different properties of counterfactual instances, such as the prediction of the desired class or similarity to the original instance. The loss function is then optimized over the training data to find the most suitable counterfactual instance. While the exact design of the loss function and the optimization algorithm vary between approaches, the high-level approach often takes the same form. 

In this work, we challenge the definition of counterfactuals inherited from supervised learning for explaining RL agents.
We examine the similarities and differences between the supervised and RL from the perspective of counterfactual explanations and argue that the same definition of counterfactual explanations cannot be directly translated from supervised to RL. Even though the two learning paradigms share similarities, in this work we demonstrate that the sequential nature of RL tasks, as well as the agent's goals, plans, and motivations, make these two approaches substantially different from the perspective of counterfactual explanations. We start by reviewing the existing state-of-the-art methods for generating counterfactual explanations in supervised learning. Furthermore, we identify the main differences between supervised and reinforcement learning from the perspective of counterfactual explanations and redefine them for RL use. Finally, we identify research questions that need to be answered before  counterfactual explanation methods can be applied
to RL and propose potential solutions.

Previous surveys of XRL recognize counterfactual explanations as an important method but do not offer an in-depth review of methods for generating this type of explanation \cite{puiutta2020explainable,heuillet2021explainability,wells2021explainable}. Previous surveys of counterfactual explanations, however, focus only on methods for explaining supervised learning models and offer a theoretical background and review of state-of-the-art approaches \cite{stepin2021survey,verma2020counterfactual,sokol2019counterfactual}. Similarly, \citet{guidotti2022counterfactual} review counterfactuals for supervised learning and offer a demonstration and comparison of different approaches. On the other hand, in this work, we focus specifically on counterfactual explanations from the perspective of RL. Additionally, while previous work has explored differences between supervised and RL for causal explanations \cite{broadXAI}, we utilize this to redefine counterfactual explanations for RL use, as well as explore challenges of applying supervised learning methods for generating counterfactual explanations directly in RL. 

The rest of the work is organized as follows. Section \ref{XRL} provides a taxonomy and a short overview of methods for explaining the behavior of RL agents. In Section \ref{vs} we identify the key similarities and differences between supervised and reinforcement learning from the perspective of explainability. Properties of counterfactual explanations are explored in Section \ref{CFs}.  Furthermore, Section \ref{Methods} offers a review of the state-of-the-art methods for generating counterfactual explanations in supervised and reinforcement learning. Finally, Section \ref{CFRL} focuses on redefining counterfactual explanations for RL and identifying challenges and open questions in this field.

\section{Explainable Reinforcement Learning (XRL)}

\label{XRL}

Recent years have seen a rise in the development of explainable methods for RL tasks \cite{puiutta2020explainable}. In this section, we provide an overview of RL framework, XRL taxonomy and offer a condensed review of current state-of-the-art XRL methods.

\subsection{Reinforcement Learning (RL)}

Reinforcement learning (RL) \cite{sutton2018reinforcement} is a trial-and-error learning paradigm for navigating sequential decision-making tasks. RL problems are usually presented in the form of Markov Decision Process (MDP) $M = (S, A, P, R, \gamma)$, where $S$ is a set of states an agent can observe and $A$ a set of actions it can perform. Transition function $P: S \times A \rightarrow S$ defines how actions change the current state and produce a new state. Actions can elicit rewards $r \in R$, whose purpose is to guide the agent toward the desired behavior. Parameter $\gamma$ is the discount factor. Agent's performance, starting from time step $t$, can be calculated as the \textit{return}:

\begin{equation}
    G_t = \sum_{i=t+1}^{T} \gamma^{i - t - 1} \cdot r_i
\end{equation} where $\gamma$ balances between the importance of short and long-term rewards. 

The goal of the RL agent is to learn a policy $\pi: S \rightarrow A$, which maps the agent's states to optimal actions. One of the most notable approaches for learning the optimal policy is deep reinforcement learning (DRL). DRL is a field at the intersection of deep learning and RL. DRL algorithms use neural networks to represent an agent's policy. This allows easier processing of unstructured observations, such as images, and overcomes scalability limitations of tabular methods such as Q-learning. DRL algorithms have been successfully applied to various domains, such as games \citep{vinyals2017starcraft,mnih2013playing}, autonomous driving \citep{kiran2021deep} and robotics \citep{kober2013reinforcement, peng2018deepmimic}. Despite their success, DRL algorithms cannot be safely deployed to real-life tasks before their behavior is fully verified and understood. However, since these methods rely on neural networks to learn the optimal policy, the agent's behavior is difficult to explain and the reasoning behind its decisions is not transparent. 

Additionally, there are a number of extensions to the RL framework. In multi-goal RL, for example, an agent operates in an environment with more than one goal and must navigate them successfully to finish the task. A simple taxi environment in which an agent needs to pick up and drop off passengers is an example of a multi-goal problem, with two subtasks. Similarly, multi-objective RL applies to tasks where the agent needs to balance between different, often conflicting desiderata. In autonomous driving, for example, an agent needs to consider different objectives such as speed, safety, and user comfort. 

\begin{table}[t]
    \centering
    \caption{Categorization of methods for explainable RL based on the XRL taxonomy (Section \ref{XRLTaxonomy}). Methods are classified based on their scope, reliance on the black-box model, time of explanation and intended audience.}
    \begin{adjustbox}{width=\textwidth}
    \begin{tabular}{@{}lcccccc@{}}\toprule
         \textbf{Method} & \textbf{Subcategory} & \textbf{Scope} & \textbf{Model-reliance} & \textbf{Explanation time} & \textbf{Audience} & \textbf{Publications}  \\ \toprule
         \multirow{2}*{Global surrogates} &  Decision trees & Global & Model-specific & Post-hoc & Developers/Experts & \citet{coppens2019distilling, liu2018toward} \\  \cmidrule{2-7}
         & Custom model & Global & Model-specific & Post-hoc & Developers/Experts & \citet{verma2018programmatically,verma2019verifiable} \\ \midrule
        \multirow{2}*{Saliency maps} & Gradient-based & Local & Model-specific & Post-hoc & Developers/Experts & \citet{wang2016dueling} \\ \cmidrule{2-7}
        & Perturbation-based & Local & Model-agnostic & Post-hoc & Developers/Experts & \citet{greydanus2018visualizing,puri2019explain}\\ \midrule
        Summaries & & Global & Model-agnostic & Post-hoc & All & \citet{amir2018highlights,sequeira2020interestingness} \\ \midrule
        \multirow{2}*{Contrastive Expl.} & Contrasting outcomes & Local & Model-agnostic & Post-hoc & All & \citet{madumal2020explainable,van2018contrastive-exp} \\ \cmidrule{2-7}
        & Contrasting objectives &   Local/Global & Model-agnostic & Post-hoc & All & \citet{juozapaitis2019explainable,sukkerd2020tradeoff,gajcin2021contrastive} \\ \midrule
        Hierarchical RL & & Global & Model-specific& Interpretable & Developers/Experts & \citet{beyret2019dot} \\ \midrule
        Visualization & t-SNE & Global & Model-agnostic & Post-hoc & Developers/Experts & \citet{zrihem2016visualizing,mnih2013playing} \\ \midrule
        
        \multirow{3}*{Other} & Abstract graphs & Global & Model-specific & Interpretable & Developers/Experts & \citet{topin2019generation} \\ \cmidrule{2-7}
        
        & Symbolic policies & Global & Model-specific & Interpretable & Developers/Experts  & \citet{landajuela2021discovering} \\ \cmidrule{2-7}
        
        & Causal confusion & Global & Model-agnostic & Post-hoc & Developers & \citet{gajcin2022reccover} \\ \bottomrule 
    \end{tabular}
        \end{adjustbox}
    
    \label{XRLtable}
\end{table}

\subsection{XRL Taxonomy}
\label{XRLTaxonomy}

According to a previous survey of XRL by \citet{puiutta2020explainable} the taxonomy of XRL methods largely overlaps with the taxonomy of XAI approaches, and classifies methods based on the \textit{reliance of the black-box model, time of interpretation, scope and target audience}.

Depending on their reliance on the black-box model, XRL methods can be \textit{model-agnostic} or \textit{model-specific}. Model-specific explanations are developed for a particular type of black-box algorithm, while model-agnostic approaches can explain any method. 

An XRL model can be either \textit{intrinsically} or \textit{post-hoc explainable.} An intrinsically explainable model has an interpretable structure, such as linear models or decision trees. Just by observing the structure of such a model, we can understand the relationships between the individual features and the outcome.  On the other hand, post-hoc explainability assumes the model is a black box and explanations are generated after the model has been trained.

According to their scope, explanations can be \textit{local} or \textit{global}. Local methods focus on interpreting a single decision in a specific instance, while global approaches explain the entire behavior of a model. For example, saliency maps are a local approach that explains an action of RL agent by highlighting parts of the state that contributed to that action. On the other hand, global surrogates use interpretable models, such as decision trees to distill a black-box policy into a more understandable format.

Additionally, we can also make a distinction between explainability methods based on their target audience. The majority of current XRL approaches focus on explaining agent behavior to developers or experts. This means that they often offer low-level, technical explanations which may be incomprehensible to a non-expert user. User-friendly explanations, on the other hand, must deal in high-level terms. They usually explain an event by comparing it to a similar event with a different outcome \cite{miller2019explanation}, or offering actionable advice on how to change the input in order to obtain a desired output. 

\subsection{XRL Methods}

In this section, we provide a brief overview of current methods in XRL. Specifically, we review some notable approaches for global and local explanation. In the category of global approaches, we review global surrogates as a notable model-specific approach and summaries as the model-agnostic approach. In local explanations, we review saliency maps, which can be both model-specific and model-agnostic, and contrastive explanations which are often model-agnostic. While most of XRL methods fall within these four categories, there are some other approaches like hierarchical interpretable RL \cite{beyret2019dot} or visualization approaches \cite{zrihem2016visualizing,mnih2013playing} which we omit from this survey. An overview of the most notable XRL methods is given in Table \ref{XRLtable}.

\subsubsection{Global Surrogates}

One of the main approaches to global explanations in RL is representing policies using an intrinsically interpretable model. This often consists of training a black-box DRL model and transforming it into an interpretable AI structure, while retaining performance. Most often, an RL policy is distilled into a tree structure, although several works introduce novel interpretable frameworks for representing RL policy.

\citet{coppens2019distilling} employ a Soft Decision Tree (SDT) to predict actions of the DRL policy. SDT is a binary tree with predetermined depth, where each node $i$ consists of a perceptron with weight $w_i$ and bias parameter $b_i$. SDT is trained to predict the optimal action from state features using a data set of state-action pairs $(s, \pi(s))$ extracted from DRL policy's $\pi$ interactions with the environment. Explanation for an individual decision is obtained by traversing the trained SDT and observing which spatial features were considered on the decision path. Even though SDT attempts to mimic the DRL policy, its performance depends on the tree depth and fails to replicate that of the black-box model. Increasing the depth of the tree leads to a larger expected reward in Mario AI benchmark task \citep{karakovskiy2012mario}, but in turn, decreases interpretability since decision paths become longer and more difficult to understand. Similarly, \citet{liu2018toward} propose a regression tree with linear models in leaves, named Linear Model U-Tree (LMUT). In contrast to \citet{coppens2019distilling}, LMUT is not used to predict the agent's next action, but rather represents the black-box policy implicitly, by approximating its Q function. Authors additionally propose an online learning algorithm that combines active interactions with the environment and learning of the Q-function directly from the black-box model. Instead of using state-action pairs as training data, a sequential data set of interactions $(s_t, a_t, r_t, s_{t+1}, Q(s_t, a_t))$ is gathered from interactions with the environment. 

In contrast to tree-based approaches, \citet{verma2018programmatically, verma2019verifiable} do not use a known structure, but introduce an interpretable RL-specific programming language. Each policy is represented as a program in a high-level, domain-specific language. \citet{verma2018programmatically, verma2019verifiable} also propose NDPS (Neurally Directed Program Search), a method for searching the program space. NDPS uses a trained black-box policy as an oracle, and attempts to find an interpretable program that approximates this policy best. NDPS consists of iterative updates of the current estimate $e$ of the oracle policy $e_N$. In each iteration, a neighborhood of policies structurally similar to the current estimate $e$ is tested, and the one most similar to oracle is promoted to the current estimate. The iterative process converges after no change to the policy has been made for several iterations. The interpretable policy is evaluated in a car-racing environment, and achieves comparable performance to the black-box DRL policy. 

Global surrogate models provide an alternative to black-box solutions, but their capabilities are usually limited by the trade-off between interpretability and performance. Moreover, they require additional time and resources for training a separate interpretable policy alongside the original, black-box one. 

\subsubsection{Summaries}
\label{summaries}
One of the most notable model-agnostic global methods is summary generation. Summaries are usually directed towards non-expert users of the system and aim to showcase specific parts of the agent's behavior to give the user an idea about the agent's capabilities and limitations. Summaries can also be beneficial to experts and developers as they can emphasize situations where an agent exhibits suboptimal behavior.

\citet{amir2018highlights} proposed HIGHLIGHTS, a method for selecting the most important states to present to the user. A state is considered to be important if making a wrong decision in the state leads to a significant decrease in expected reward. Formally, the importance of a state $s$ is calculated as the difference between the largest and the smallest expected reward, attainable from $s$ depending on the chosen action:

\begin{equation}
    \mathcal{I}(s) = \max_a Q(s, a) - \min_a Q(s, a)
\end{equation}

A limited number of transitions relating to states with the highest importance score are presented to the user in the form of a summary. As the authors too acknowledge, this approach comes with significant drawbacks, such as sensitivity to the number of actions. \citet{sequeira2020interestingness} addressed some limitations and expanded the HIGHLIGHTS algorithm by proposing four heuristics that determine whether a state is interesting enough to be included in the summary:

\begin{itemize}
    \item \textit{Frequency analysis} detects situations that are frequently or infrequently visited by the agent. Frequency can be calculated for states: $n(s)$, state-action pairs: $n(s, a)$ or transitions: $n(s, a, s')$.
    \item \textit{Execution certainty} refers to the agent's confidence in its decision in a specific state. Situations of high confidence could indicate the agent's capabilities, while those in which the agent is indecisive could showcase its limitations.
    \item \textit{Transition value} is a measure of \textit{favorableness} of the situation. Favorable states are those whose state value $V(s)$ is higher than for states in its immediate surroundings, while unfavorable states see the local minima of $V(s)$.
    \item \textit{Sequence analysis} uncovers most likely paths from interesting states (extracted according to the previous three conditions) to local maxima states in terms of the state value function.  
\end{itemize}

Summary methods provide a way of condensing the agent's behavior and locating situations where the agent acts in an unexpected or otherwise interesting way. At the moment, however, these methods rely on heuristics to determine whether a state is important enough to be included in the summary. Whether the state will be included in the summary only depends on its basic features, such as visitation frequency and action certainty, or its expected reward. Other situations of interest, such as states in which the agent's goals change, or where an unexpected outside event impacts its plan are not explored. However, in a dynamic environment, understanding how the agent adjusts its goals and plans might be necessary to fully grasp the agent's capabilities and limitations. Current methods also require access to the agent's state and Q value functions, which can be inexact as they are the agent's approximations of expected reward, as well as difficult to compare between agents. Additionally, while useful for visual tasks, summaries are not an appropriate technique for explaining tasks with other types of input, such as personal assistants \citep{amir2018highlights}.

\subsubsection{Saliency Maps}

Saliency maps have originally been used to interpret decisions of neural networks for supervised learning tasks \cite{simonyan2013deep}, but have also been applied as an out-of-the-box solution to the problem of explainable RL. This approach tries to explain why an action was chosen in a state by highlighting specific parts of the image that the agent focused on while making the decision. There are two approaches to generating saliency maps: \textit{perturbation-based} and \textit{gradient-based}. Perturbation-based methods distress parts of the image and compare the model's decision before and after alternation, to assess the importance of a feature. This can mean either removing, blurring or in another way altering the part of the image. Intuitively, if the agent's decision changes significantly after altering a feature, then the feature is important for the prediction. Gradient-based methods, however, rely on calculating the Jacobian of the neural network with respect to the input in order to see how much specific parts of the image impacted the decision. Gradient-based methods require access to the model's internal parameters and structure, while perturbation methods use the model as an ``oracle'' and only require access to the model's output. 

\begin{figure}[t]
    \centering
    \includegraphics[width=0.3\textwidth]{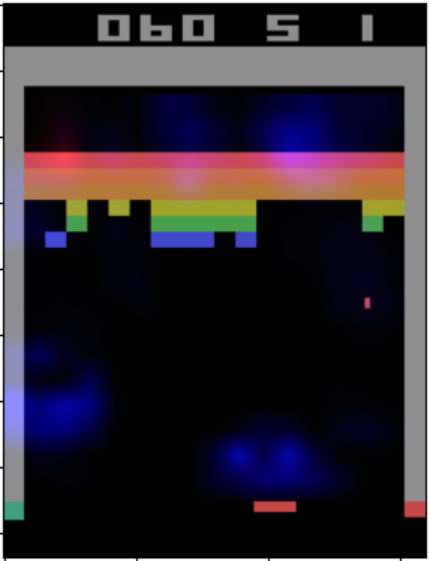}
    \quad
    \includegraphics[width=0.3\textwidth]{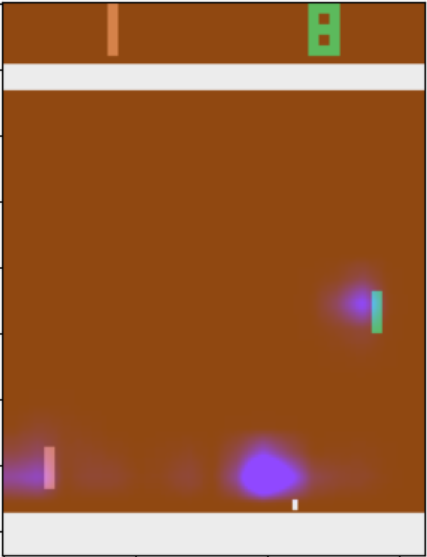}
    \quad
    \includegraphics[width=0.3\textwidth]{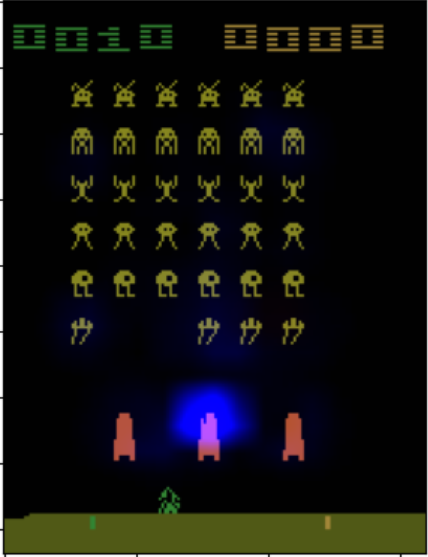}
    \caption{Saliency maps for explaining decision of Atari agents for \textit{Breakout}, \textit{Pong} and \textit{Space Invaders} environments , obtained with algorithm provided in \citet{greydanus2018visualizing}. When making the decision in the pictured states, the agent focuses on the highlighted areas in the image.}
    \label{fig:my_label}
\end{figure}

\citet{greydanus2018visualizing} used perturbation-based saliency maps to explain agents' actions in Atari games. To determine their contribution to a specific action, Gaussian blur was added to the image features. The approach was used to expose how agents learn interesting strategies, such as ``tunneling'' in \textit{Breakout} environment. Tunneling refers to a situation where an agent aims the ball through the brick wall to create a tunnel and then propels the ball to the top of the screen allowing it to bounce between the ceiling and the brick wall, collecting dense rewards. Saliency maps uncovered that while creating the tunnel the agent attends to the top of a brick wall, where its future rewards lie. Additionally, \citet{greydanus2018visualizing} apply saliency maps to uncover overfitting, in a situation where an agent learns to rely on a feature that is correlated with the optimal action but does not cause it. The overfitting scenario is created by adding hint pixels, representing the most likely expert action, to the raw Atari state. Although the overfitting agent shows good performance, saliency maps uncover that it bases its decisions on the added hint feature. Finally, \citet{greydanus2018visualizing} also showcased the potential benefits of saliency maps for debugging. After observing the suboptimal behavior of the agent in Ms Packman environment, the authors apply saliency maps to uncover that the agent was not attending to the ghosts, resulting in the failure to avoid them. \citet{greydanus2018visualizing} recognize that capturing this kind of faulty behavior can inspire developers to reexamine the reward function that the agent is following. Similarly, while \citet{puri2019explain} proposed SARFA (Specific and Relevant Feature Attribution), a method for generating perturbation-based saliency maps, and applied it to explain the moves of a chess agent. In this work, perturbation is performed by removing the piece from the chessboard. The action is further explained by comparing the difference in the expected reward of taking the action in the original and perturbed state. SARFA addresses two limitations of previous perturbation-based methods -- \textit{specificity} and \textit{relevance}, and uses their harmonic mean to calculate feature saliency. Specificity means that the saliency of a feature should only focus on the action being explained. Perturbation of a feature should only affect the saliency of that feature if it significantly affects the action being explained, while not impacting other actions. Relevance addresses the need for saliency of a feature to only include those perturbations whose effects are relevant to the action being explained. In other words, perturbations should have a significant effect on the action in question, without impacting other actions, for the feature to be considered important.

Gradient-based saliency maps have not yet found broad applications in the field of RL. Few works utilize saliency maps to demonstrate and evaluate the behavior of newly proposed methods. \citet{wang2016dueling} introduce Dueling Networks which separately estimate the state-value function and the advantage of each action, by implementing two different streams in the DNN structure. Dueling networks can distinguish between states in which making a good decision is crucial and those where there is no significant difference between actions. Gradient-based saliency maps are used to demonstrate and validate the behavior of dueling networks in a driving environment, showing that in critical situations advantage stream attends to nearby cars, while if the road is clear it does not attend to any features, as its decision is not important.

Saliency maps are a useful method for the visualization of important areas in the state space. However, they explain the action only as a function of input features. As such, this approach might fit well into the single-step decision-making process of supervised learning algorithms, where a decision can be easily interpreted in terms of individual feature contributions. In RL, however, saliency maps focus only on the current state to explain a decision, while ignoring other possible causes of behavior such as the agent's plan, current goal, or outside events.

\subsubsection{Contrastive Explanations}

Contrastive explanations explain why an agent chose a specific behavior over another. Evidence from social sciences suggests that humans prefer explanations that contrast different alternative events or options \citep{miller2019explanation}. According to \citet{lipton1990contrastive}, contrasting explanations can help narrow down the list of causes of an event. Naturally, humans prefer explanations that are selective and present a few most important causes, instead of being overwhelmed by a substantial list of all potential causes of an event \citep{molnar2019}. All of this makes contrastive explanations user-friendly.

The majority of contrastive explanations in RL focus on comparing the outcomes of actions $a$ and $a'$, to uncover why the agent chose $a$. For example, \citet{van2018contrastive-exp} generate contrastive explanations by comparing the expected outcomes of two actions in a specific state. Firstly, the agent's policy is made more understandable by mapping states and outcomes to a hand-crafted set of interpretable concepts. Outputs describe the agent's position in understandable terms -- in the grid-world environment, for example, the agent can be at the goal, in a trap, or near a forest. Using hand-crafted states and outputs, the approach can explain why the agent chose one or more consecutive actions of the fact policy $\pi_t$ and did not follow the alternative foil $\pi_f$. Both policies are unrolled in the environment, starting in state $s$ for a set number of steps $n$ and their visited paths $Path(s, \pi_t)$ and $Path(s, \pi_f)$ are recorded. Contrastive explanation can then be generated by comparing the two paths. Authors propose taking a set difference $Path(s_t, \pi_t) \setminus Path(s_t, \pi_f)$ to obtain outcomes unique to $\pi_t$, or calculating a symmetrical difference $Path(s_t, \pi_t) \Delta Path(s_t, \pi_f)$.

Similarly, \citet{madumal2020explainable} answer the question \textit{``Why did agent take action $a$ instead of $b$ in state $s$?''} by comparing the outcomes of two potential actions. Instead of unrolling the alternative policies in the environment, however, authors use a causal model of the environment named \textit{action-influence model}. In the action-influence model, nodes represent state features, and arcs indicate causal relationships. Each arc is also assigned a structural equation that defines the nature and strength of the causal effect. Additionally, each arc is assigned an action, representing the causal effect after the action is performed. It also allows for a special type of node called the reward node, to represent the effect of actions and features on the reward. The authors use a hand-crafted causal model and learn structural equations from transition data using linear regression to represent the action-influence model. To generate contrasting explanations, causal chains of actions $a$ and $b$ are compared. To reduce explanation complexity, the authors propose that only minimal chains are used to represent the causal effects of an action. For a complete chain, the minimal chain consists of only the first and last arc including their source and destination nodes, ignoring the intermediate results. The final contrastive explanation is obtained by comparing differences between minimal causal chains of actions $a$ and $b$.

In multi-objective RL, where an agent needs to balance between different, often conflicting goals, contrastive explanations focus on trade-offs between objectives for different actions. \citet{sukkerd2020tradeoff} propose a method for generating contrasting explanations that show how the trade-off between different objectives contributed to the choice of the action. Explanation is presented in terms of quality attributes (QA), which represent specific objectives (e.g. execution time, energy consumption...). To provide contrasting explanations, the authors propose comparing the policy to a subset of Pareto-optimal policies. The resulting contrastive explanations show the trade-offs in QAs that guided the choice of action:

\begin{quote}
    I could improve on $QA_i$ following policy $\pi_i$. However, this would worsen the $QA_j$ by an amount $qa_j$. I chose the action because improvement in $QA_i$ is not worth the deterioration of $QA_j$
\end{quote}

Similarly, \citet{juozapaitis2019explainable} convert the single-objective problem into a multi-objective one, by decomposing the reward into semantically meaningful components. Furthermore, they explain the difference between two actions in one state in terms of reward components. Formally, authors split the reward into a reward vector $\vec{R}: S\times A \rightarrow \mathbb{R}^{|C|}$ where $C$ represents a set of different reward types. Each component $R_c(s, a)$ of the reward vector corresponds to the specific reward type, and the overall reward can be obtained as the sum of individual rewards: $R(s, a) = \sum_{c \in C} R_c(s, a)$. Similarly, the Q-value can be also split into individual components $Q_c^\pi$, where each component corresponds to one reward type. The original Q-value can then be calculated as a sum of individual reward-specific values: $Q^\pi = \sum_{c \in C} Q_c^\pi$. Authors then propose a novel learning algorithm $drQ$, inspired by Q-learning, which supports individual rewards and updates separately to each $Q_c^\pi$. The difference between two actions $a_1$ and $a_2$ in state $s$ is then explained by analyzing the trade-off between the reward components. In order not to overwhelm the user, authors propose only presenting them with minimal contrastive explanations, which contain only critical positive and negative reasons for a decision. 

On the global level, \citet{gajcin2021contrastive} propose a post-hoc summarization method for contrasting the behavior of two agents. The approach focuses on the difference in preferences between two policies trained on different reward functions. The approach aims to compare and summarize the differences between two policies, by focusing only on the differences in their preferences and ignoring potential differences in their abilities. To explore differences between policies, authors first unroll the policies in the environment and record disagreement states -- states in which policies disagree on the best action choice. Both policies are also separately unrolled from each disagreement state for a set number of steps and recorded as disagreement trajectory pairs. To filter out disagreements that stem from the difference in preferences from those that stem from different abilities the authors propose three conditions. The disagreement between two policies is preference-based if:

\begin{enumerate}

    \item  \textit{Both policies are highly confident in their decision in the disagreement state.}
    \item  \textit{Both policies have similar evaluations of the disagreement state $s_d$.}
    \textit{To estimate this similarity we evaluate the expression}
    \item \textit{After unrolling policies in the environment for $k$ steps from the disagreement state, both policies have similar evaluations of their outcomes.}
\end{enumerate}

Intuitively, the conditions describe disagreement where both policies are sure of their decision, have similar evaluations of their current state, and see the same potential in the environment, but differ in their preferred path to achieving that potential. Preference-based disagreement data is then used to generate contrasting explanations about the policies. For example, after comparing safety-based and speed-based driving policies $\pi_A$ and $\pi_B$, the approach generates the following explanation:

\begin{displayquote}
    \textit{\textit{Policy $\pi_A$ prefers states with $x$ smaller, $y$ smaller, $h$ smaller, $v$ smaller, $u$ larger
compared to policy $\pi_B$}.}
\end{displayquote}

where state features $x$ and $y$ denote car's location, $h$ is the heading, $v$ velocity and $u$ steering wheel angle.

Contrastive explanations are a user-friendly type of explanations that can summarize the most important causes of a decision. However, current methods mostly focus on comparing state-based outcomes of decisions. There is room for extending contrastive methods to include goals, objectives, or unexpected events to explain a preference for one action over another. These explanations could be especially useful in multi-goal, multi-objective, or stochastic RL environments. 

\begin{figure}[t]
    \centering
    \includegraphics[width=0.45\textwidth]{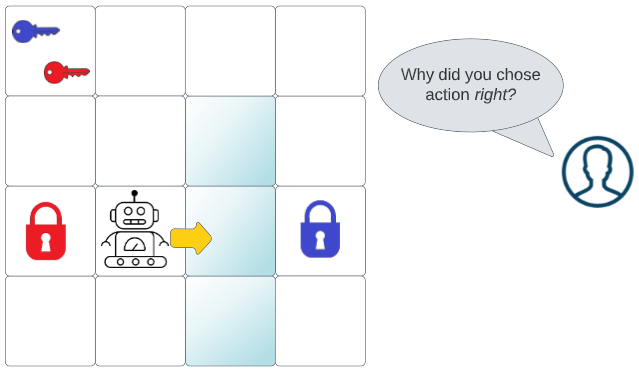}\\
    
    \begin{subfigure}[b]{0.3\textwidth}
         \centering
         \includegraphics[width=1\textwidth]{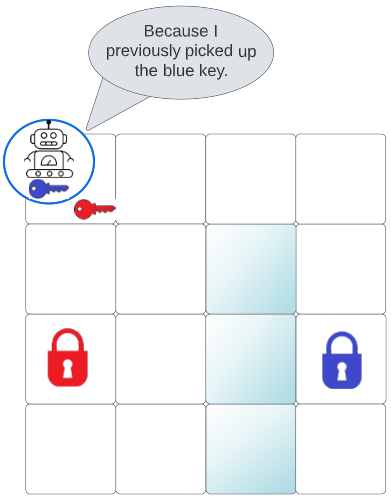}
         \caption{}
         \label{a}
     \end{subfigure}
     \hfill
     \begin{subfigure}[b]{0.362\textwidth}
         \centering
         \includegraphics[width=\textwidth]{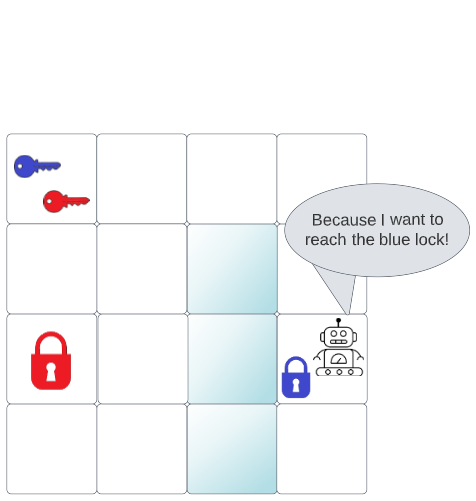}
         \caption{}
         \label{b}
     \end{subfigure}
     \hfill
     \begin{subfigure}[b]{0.287\textwidth}
         \centering
         \includegraphics[width=\textwidth]{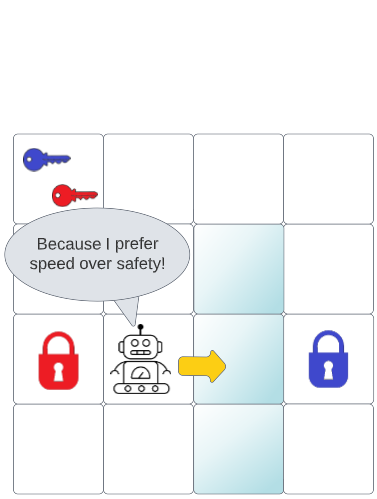}
         \caption{}
         \label{c}
     \end{subfigure}
     
    \caption{Example of different types of explanations in a simple RL gridworld environment. Agent's task is to pick up a key of any color, navigate to the lock of the same color, and open it. Blue squares are ice, and stepping on them brings a large penalty to the agent with some small probability. Top: The user asks the agent why it chose action \textit{right} in a specific state. Bottom left: The agent's explanation refers to the previously visited state where it picked up the blue key. Bottom middle: The agent explains its decision with a temporally distant goal. Bottom right: The agent explains its choice by expressing its preference between two conflicting objectives.}
    \label{slvsrl}
\end{figure}

\section{Explainability in Supervised vs. Reinforcement Learning: Similarities and Differences}
\label{vs}

Supervised and reinforcement learning are often used to solve different types of tasks. For example, supervised learning is used for learning patterns in large amounts of labeled data, while RL focuses on sequential tasks, and learns them from scratch through interactions with the environment. However, from the perspective of explainability, the two approaches share notable similarities. Most importantly, the lack of transparency in supervised and reinforcement learning stems from the same source -- the underlying black-box model. In supervised learning, a black-box model is used to map input states to their labels. Similarly, in RL tasks, the model maps the agent's state to an optimal action. Most explainability methods focus on deep learning and DRL scenarios, where the underlying black-box model is a neural network \cite{vouros2022explainable}. The high-level approach of the two paradigms is in fact identical -- a black-box model takes in often a multi-dimensional input consisting of features and produces a single output. For this reason, a model-agnostic approach trying to explain the model's prediction in a supervised learning task can easily be applied to an RL task for explaining a choice of action in a specific state. For example, saliency maps have been successfully repurposed for RL \cite{huber2021benchmarking}, despite being developed with supervised learning in mind \cite{simonyan2013deep}. Other local explanation methods such as LIME \cite{ribeiro2016should} and SHAP \cite{lundberg2017unified} have also been used to interpret the effect of individual features on the action in the state \cite{dethise2019cracking,zhang2020explainable}. 

However, despite their similarities, supervised learning, and RL frameworks differ in a few notable points and thus often require different explanation techniques. In Figure \ref{slvsrl} we present some illustrations of the main differences between supervised and RL from the perspective of explainability:

\begin{enumerate}
    \item \textit{State independence:} one of the main assumptions of supervised learning algorithms is the independence of instances in the training data. RL, however, focuses on sequential tasks where states are temporally connected. There is an inherent causal structure between visited states and chosen actions during the agent's execution. A certain state is only visited because previous states and actions led to it \cite{broadXAI}. The causal links can be important components to explain the outcomes of the RL agent's behavior. Figure \ref{a} illustrates using previously visited states to explain the current decision. The agent justifies its decision of choosing the action \textit{right} by remembering that it had previously picked up the blue key and should navigate towards the blue goal.

    \item \textit{Sequential execution:} supervised learning approaches are limited to one-step prediction tasks, while RL framework specializes in sequential problems. While methods for explaining supervised learning models only need to rely on model parameters and input features to explain a prediction, the causes of the RL agent's decisions can be in its past or the future \cite{broadXAI}. For this reason, explanations of decisions in RL cannot be contained to the current time step but may need to include temporally distant causes. Figure \ref{b} shows how temporally distant states can motivate an agent's decision.  
    
    \item \textit{Explanation components}: in supervised learning, a model's decision is explained as a function of input features and model parameters. In RL, however, an agent's decision-making process can be more complex and can include plans, goals or unexpected events. To fully understand the behavior of RL agents, explanations often cannot be limited to only state features, but should also include other possible causes of decisions. For example, in Figure \ref{c} an agent can explain its decision by comparing two different objectives it balances between. Even though an agent is choosing the fastest way to the goal, stepping on the ice can lead to a large penalty. To understand why an agent would choose a potentially dangerous decision, it is necessary to know that it prefers speed over safety. 
    
    \item \textit{Training data set:} while a necessary component in supervised learning approaches, it does not have a natural equivalent in an RL framework. For this reason, explainability methods for supervised learning which rely on training data sets (e.g., LIME, counterfactuals) might have to be adjusted to be applicable to RL tasks.

\end{enumerate}

Supervised learning explainability approaches are limited to one-step prediction tasks, and as such have limited applications in RL. While they can be used to explain a decision based on the current state features, an RL agent's decisions are often influenced by a wider range of causes and require richer, temporally extended explanations. For that reason, there is a need for RL-specific explainability techniques that account for the particularities of the framework.

\section{Counterfactual Explanations}
\label{CFs}

At the moment, a substantial amount of research in XAI is focused on developing explanations for experts familiar with AI systems. However, from the perspective of non-expert users, such explanations might be too detailed and difficult to understand. End-users are more likely to be interested in more abstract explanations of the system \citep{dazeley2021levels}. For example, consider the 2018 incident when an Uber self-driving vehicle accidentally killed a pedestrian after failing to avoid them while crossing the road \citep{knight_2018}. In this situation, developers and experts might be interested in uncovering which parameters or input features contributed to the fatal failure, in order to be able to repair it. On the other hand, a non-expert user is more likely to be interested in high-level questions, such as \textit{``Did the car not recognize the person in front of it?''} or \textit{``Did the car believe it had the right of way?''} \citep{dazeley2021levels}. Although the research in developing low-level, expert-oriented explanations has gained momentum, user-friendly explanation methods are less explored. However, developing user-friendly explanations is necessary in order to build trust and encourage user interactions with the black-box systems. 

In this work, we explore one of the most notable examples of user-friendly explanations -- \textit{counterfactual explanations}. Counterfactuals are local explanations that attempt to discover the causes of an event by answering the question: \textit{``Given that input $x$ has been classified as $y$, what is the smallest change to $x$ that would cause it to be classified as an alternative class $y'$?''}. For example, counterfactuals can be applied to answer questions such as \textit{``Since my application was rejected, what are some examples of successful loan applications similar to mine?''} or \textit{``How can I change my application for it to be accepted in the future?''}. The explanation is usually given in the form of a \textit{counterfactual instance} -- an instance $x'$ as similar as possible to the original $x$, but producing an alternative outcome $y'$. Formally, according to \citet{wachter2017counterfactual}, counterfactual explanations can be defined as statements:

\begin{table}[t]
    \caption{Some important terms in counterfactual explanations research and the notation used in this work.}
    \centering
    \begin{adjustbox}{width=\textwidth}
        \begin{tabular}{@{}lcc@{}}
        \toprule
        \textbf{Term} & \textbf{Symbol} & \textbf{Meaning} \\ \midrule
        Black-box model & $f$ & Model being explained \\
        
        Original instance & $x$ & Instance being explained \\
        
        Original output & $y$  & Output produced for $x$ by $f$; $y = f(x)$\\
        
        Counterfactual output & $y'$ & Desired output; $y' \neq y$  \\
        
        Counterfactual instance & $x'$ & Instance similar to $x$, but producing counterfactual output $y'$ \\
        
        Counterfactual perturbation & $\delta$ & Perturbation that needs to be added to $x$ to obtain $x'$; $x' = x + \delta$\\
        
        \bottomrule
        \end{tabular}
    \end{adjustbox}
    \label{cf}
\end{table}

\begin{quotation}

Score y was returned because variables V had values (v1, v2, . . .) associated with them. If V instead had values (v1', v2', . . .), and all other variables had remained constant, score y' would have been returned.
\end{quotation}

The statement consists of two main factors: \textit{variables} that the user can alter, and the \textit{score}. In supervised learning, variables represent input features that a black-box model uses to make a prediction, for example, pixels in image-based tasks. The prediction of the black-box model represents the score. The alternative outcome $y'$ can either be specified or omitted, in which case the counterfactual can be an instance producing any outcome other than $y$. Current methods either directly search for a counterfactual instance $x'$ similar to $x$, or a small perturbation $\delta$ which can be added to $x$ to achieve a desired outcome, in which case $x' = x + \delta$. A summary of the terminology used throughout the work is given in Table \ref{cf}.

The following properties make counterfactuals user-friendly explanations:

\begin{enumerate}
    \item \textit{Actionable}: counterfactual explanations give the user actionable advice, by identifying which parts of the input should be changed in order for the model's output to change. Users can use this advice to alter their input and obtain the desired outcome.
    \item \textit{Contrastive}: counterfactual explanations compare real and imagined worlds. Contrastive explanations have been shown to be the preferred way for humans to understand the causes of events \cite{miller2019explanation}.
    \item \textit{Selective}: counterfactual explanations are often optimized to change as few features as possible, making it easier for users to implement the changes. This also corresponds to the human preference for shorter rather than longer explanations \cite{molnar2019}.
    \item \textit{Causal}: counterfactuals correspond to the third tier of Pearl’s Ladder of Causation \cite{pearl2018book} (Table \ref{pearl-hierarchy}). The first rung of the hierarchy corresponds to associative reasoning and answers the question
    “If I observe X what is the probability of Y occurring?”. This rung also aligns
    with statistical reasoning, as the probability in question can be estimated from data. The second rung requires interventions to answer the question “How would Y
    change if I changed X?”. Finally, the third rung addresses the question “Was
    it X that caused Y?”. Answering this question requires imagining alternative
    words where X did not happen and estimating whether Y would occur in such
    circumstances.
    \item \textit{Inherent to human reasoning}: Humans rely on generating counterfactuals in their everyday lives. By imagining alternative worlds, humans learn about cause-effect relationships between events. For example, by considering the counterfactual claim \textit{``Had the car driven slower, it would have avoided the accident''}, the relationship between the car speed and the accident can be deduced. Additionally, humans use counterfactuals to assign blame or fault for a negative event \cite{byrne2019counterfactuals}. 
\end{enumerate}

As user-friendly explanations, counterfactuals are important for ensuring the trust and collaboration of non-expert users of the system. Since they offer the user actionable advice and can be possibly used to instruct users how to change their real-life circumstances, counterfactual explanations can severely influence user's trust. While providing the user with useful counterfactuals can help the user better navigate the AI system, offering the user a counterfactual that proposes unrealistic changes or does not lead to desired outcomes can be costly and frustrating for the user, and further erode their trust in the system. For this reason, it is necessary to provide a way to ensure usefulness and evaluate the quality of counterfactual explanations.

\begin{table}[t]
    \centering
    \caption{Pearl's Ladder of Causation \cite{pearl2018book}}
    \begin{adjustbox}{width=1\linewidth}
        \begin{tabular}{@{}lccc@{}}
        \toprule
            \textbf{Rung} & \textbf{Activity} & \textbf{Question} & \textbf{Example} \\ \midrule
            Association & Observing & What is the probability of Y if I observed X? & What does a symptom tell us about the disease? \\
            Intervention & Intervening & What will Y be if I change X? & If I take aspirin, will my headache be cured?  \\
            Counterfactual & Imagining & Was it X that caused Y? & Was it aspirin that cured my headache?\\
            \bottomrule
        \end{tabular}
    \end{adjustbox}
    \label{pearl-hierarchy}
\end{table}

\subsection{Properties of Counterfactual Explanations}

Due to their potential influence on user trust, there is a need to develop metrics for evaluating the usefulness of counterfactual explanations.  Additionally, counterfactual explanations can suffer from the Rashomon effect -- it is possible to generate a large number of suitable counterfactuals of a specific instance \cite{molnar2019}. In this case, further analysis and comparison is necessary to select those most useful. For this reason, 
multiple criteria for assessing the quality of the generated counterfactual explanation have been proposed \cite{verma2020counterfactual}:

\begin{enumerate}
    \item \textit{Validity}: the counterfactual must be assigned by the model to a different class to that of the original instance. If a specific counterfactual class $y'$ is provided, then validity is satisfied if the counterfactual is classified as $y'$.
    \item \textit{Proximity}: counterfactual $x'$ should be as similar as possible to the original instance $x$.
    \item \textit{Actionability}: the counterfactual explanation should provide users with meaningful insights into how they can change their features, in order to achieve the desired outcome. This means that suggesting that a user should change their sensitive features (e.g. race) or features that are immutable (e.g. age, country of origin) should be avoided, as it is not helpful and may be offensive. 
    \item \textit{Sparsity}: ideally, the counterfactual should require only a few features to be changed in the original instance. This corresponds to the notion of a selective explanation, which humans find easier to understand \citep{molnar2019}.
    \item \textit{Data manifold closeness}: the counterfactual instances rely on modifying existing data points, which can lead to generation of out-of-distribution samples. To be feasible for users, counterfactuals need to be realistic.
    \item \textit{Causality}: while modifying the original sample, counterfactuals should abide by the causal relationship between features. For illustration, consider an example by \citet{mahajan2019preserving} of a counterfactual which suggests that the user should increase their education to Masters, without adjusting their age. Even though such a counterfactual instance would be actionable and fall within the data manifold (as there probably are people of the same age as the user, but with a Master's education), the proposed changes would not be feasible for the current user.
    \item \textit{Recourse}: ideally, the user should be offered a set of discrete actions that can transform the original instance into a counterfactual. This property is closely related to actionability and causality -- offered sequence of actions should not change the immutable features, and should consider causal relationships between the variables, to understand how changing one feature can influence the others. The field of recourse started as separate from counterfactual explanations, but since explanation methods have become more elaborate and user-centered, this line has been blurred \cite{verma2020counterfactual}.
\end{enumerate}

\section{Counterfactual Explanations for Supervised Learning}
\label{Methods}

Current approaches for generating counterfactuals for RL are inspired by methods from supervised learning. For that reason, in this section, we review some of the most notable methods for generating counterfactuals in supervised learning, classify them based on their approach and evaluate them based on the counterfactual properties. 

\subsection{Counterfactual Search}

In one of the first approaches for generating counterfactuals, \citet{wachter2017counterfactual} proposed minimizing a two-part loss function over instances $x'$ in the data set:

\begin{equation}
    L(x, x', y, y') = \lambda(f(x') - y')^2 + d(x, x')
\end{equation}

where $f$ is the black-box model, $x$ is the original instance, $y$ is the label of the original instance and $y'$ is the desired outcome. The first term ensures that the counterfactual prediction is close to the desired class, while the second one minimizes the distance between the original and counterfactual instance. Parameter $\lambda$ determines the balance between two objective quantities. Gradient descent is used to minimize the loss function over the training data set. This approach focuses only on two criteria, validity, and proximity, and can result in counterfactuals that are unrealistic, or require a large number of features to be changed.

To address the issues of sparsity and data manifold closeness, \citet{dandl2020multi} expand the loss function with two components -- sparsity and data manifold closeness. Sparsity measures the number of features that had to be changed to generate the counterfactual, while data manifold closeness minimizes the distance between the counterfactual and the nearest observed instance. Multi-objective optimization is used, to generate counterfactuals that enable different trade-offs between objectives. Specifically, the Nondominated Sorting Genetic Algorithm (NSGA-II) \cite{deb2002fast} was applied, which selects suitable instances through repeated random recombination, mutation, and ranking of the obtained solutions. The approach generates a Pareto set of counterfactuals, representing different trade-offs between the objectives.

\citet{looveren2021interpretable} ensure that generated counterfactuals fall within the data manifold by defining prototypes of each class and directing the search towards one of these instances. To that end, the authors define a prototype loss, which measures the distance in the encoding space between the counterfactual and the representative of the target class.
The prototype loss pushes the search in the direction of the nearest prototype of a suitable class. To optimize the objective function, fast iterative shrinkage-thresholding algorithm (FISTA) \cite{beck2009fast} is applied. To ensure actionability, the approach also allows users to define immutable features. 


\begin{figure}[t]
    \centering
    \includegraphics[width=0.25\textwidth]{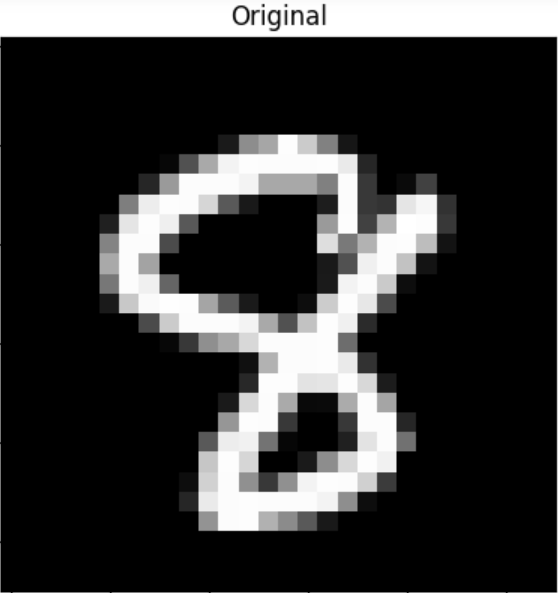}\quad
    \includegraphics[width=0.25\textwidth]{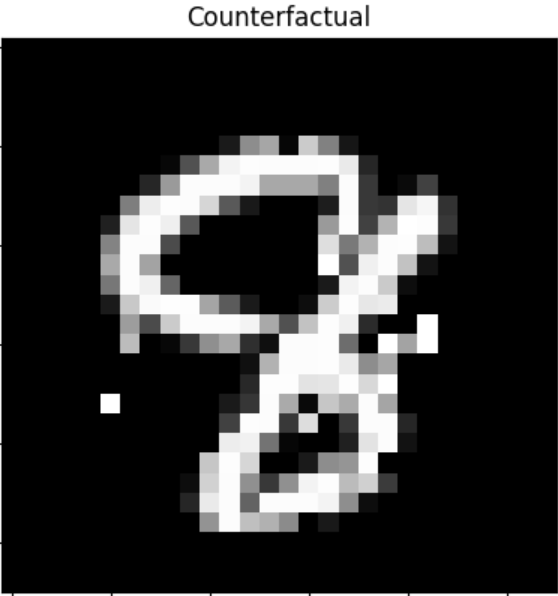}\quad
    \includegraphics[width=0.25\textwidth]{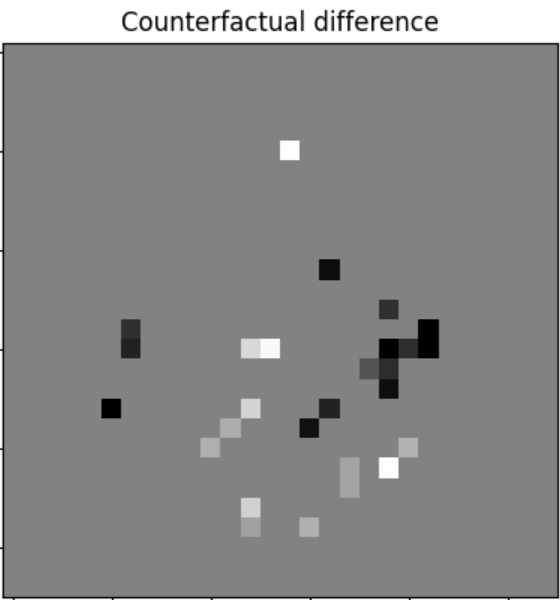}
    \caption{Counterfactual generation using growing spheres \cite{laugel2017inverse} for MNIST dataset. \textbf{Left}:
    original instance classified as 8 \textbf{Middle}: the closest counterfactual instance {classified as 9}. \textbf{Right}: pixel difference between the original and counterfactual instances.}
    \label{GrowingSpheres}
\end{figure}

Extending \citet{wachter2017counterfactual}, DICE \cite{mothilal2020explaining} generates diverse counterfactuals, to offer the user a number of possible ways to change their input. DICE introduces a diversity objective and optimizes a loss function which additionally includes proximity and validity objectives using gradient descent. To ensure actionability, users can provide constraints on immutable features. Additionally, adjustments are made to generated counterfactuals to ensure their sparsity, and each continuous feature is restored back to its value in $x$ in a greedy approach until $f(c)$ changes. DICE does not ensure that generated counterfactuals fall within the data manifold, and can result in unrealistic explanations.

\citet{laugel2017inverse} propose counterfactual search using growing spheres (Figure \ref{GrowingSpheres}). The approach minimizes the validity, proximity, and sparsity objectives. Growing spheres algorithm utilizes a two-step approach and optimizes the objectives sequentially, instead of simultaneously. The algorithm does not require access to the training data but rather relies on generating instances. To optimize the proximity objective, the algorithm iteratively generates instances in the vicinity of $x$ and examines them, until the validity constraint is met. After obtaining the closest counterfactual $x'$, its feature values are iteratively reset to their values in the original instance $x$, starting with the features with the smallest difference in values between $x$ and $x'$, ensuring sparsity. The growing spheres approach relies on generating a new instance, which can result in unrealistic counterfactuals. The approach also does not address the issue of actionability. 

\citet{1283} propose LORE, a local approach for generating counterfactuals. LORE builds a local interpretable model in the neighborhood of the instance being explained. The neighborhood is generated using a genetic algorithm. Then, factual and counterfactual rules are extracted from the interpretable model. By comparing the factual and counterfactual rules, we can deduce which features need to change and by how much to obtain the desired outcome.

To ensure the counterfactual instance falls with the data manifold, REVISE \cite{joshi2019towards}, optimizes the loss function only over likely instances. A generative model is trained to sample instances from the data distribution. Variational autoencoder (VAE) is used to represent the underlying data distribution.

Similarly, FACE \cite{poyiadzi2020face} ensures that the counterfactual is likely using a graph-based approach (Figure \ref{temporality}). Each instance in the data set is presented as a node in a directed graph, and the presence of an edge between two nodes indicates that there is a connection between them. Edges are weighted according to the density of instances between the two nodes. The shortest path algorithm is then used to find an instance closest to $x$ with a different prediction. As the graph is weighted according to data density, FACE ensures there is a dense path between the original and counterfactual instance.

\citet{mahajan2019preserving} propose a causal proximity measure, that can be included in any loss-based counterfactual method and ensures counterfactuals follow causal constraints between features. The approach assumes access to the structural causal model (SCM) of the features. A SCM is a directed graph, where each node corresponds to a feature, and arcs indicate causal relationships. Each arc is assigned a structural equation, which describes how the source node influences the sink. Causal proximity does not depend on the original instance $x$ and instead ensures that features in the counterfactual instance follow causal constraints.

ReLACE \cite{chen2021relace} reimagines the problem of finding counterfactuals as a search for a sequence of actions that lead from $x$ to $x'$. Authors suggest that this problem can be solved by learning a reinforcement learning (RL) policy which learns an optimal sequence of actions between $x$ and $x'$ In each time step, the instance $x$ is modified using an available set of actions, and the reward is assigned to the agent corresponding to how well the current state satisfies counterfactual properties of proximity and validity. The optimization is terminated when a state which satisfies the prediction constraint is obtained. To learn a policy that generates counterfactual instances a P-DQN agent is used. RELACE is model-agnostic and can quickly generate counterfactuals for each instance simultaneously. However, this approach does not address the problem of data manifold closeness of generated instances. Additionally, since each action forcibly changes feature values, causality constraints that exist between features may be violated in counterfactual instances.

Similarly, \citet{samoilescu2021model} propose using DRL algorithms to efficiently generate batches of counterfactuals in one forward pass. In contrast to RELACE \cite{chen2021relace}, the authors search for the counterfactual perturbation in the latent space instead of the high-dimensional state space. An encoder is trained to transform instances into their latent representation. To learn an RL policy that converts instances to their counterfactual counterparts, the DDPG \cite{lillicrap2015continuous} algorithm is used. The actor-network serves as a counterfactual generator, meaning that instead of outputting an action to alter the instance, the actor outputs the counterfactual itself. The proposed algorithm is model-agnostic, and can efficiently generate batches of counterfactuals. However, the authors do not explore whether the generated counterfactuals conform to the causal rules of the environment. Additionally, the issue of recourse is not considered, as the actor-network outputs counterfactuals instead of actions that could transform the original instance. 

\subsection{Algorithmic Recourse}

The field of algorithmic recourse is similar to that of counterfactual explanations, and the two are often researched and presented side by side \cite{karimi2020survey, karimi2021algorithmic}. With algorithmic recourse, the user is not only offered a favorable situation in the form of a counterfactual instance but also a set of actions that can be used to transform the original instance into a counterfactual. In this way, the user receives an actionable recommendation, which can help them change their input features and receive a different outcome from the system \cite{karimi2020survey}.

In one of the first works on recourse, \citet{ustun2019actionable} explores the problem of finding a sequence of actions that flips the decision of a linear classifier $f$. The authors assume a simple binary classifier with weights $w \in \mathbb{R}^d$, where $f(x) = sgn (<x, w>)$, and undesirable outcome $y = -1$. The authors then search for a least expensive vector of actions $a* = [0, a_1, \dots, a_d]$ that such that $f(x + a) = +1$, using integer programming.

\citet{sharma2019certifai} propose CERTIFAI, an approach that uses a genetic algorithm to alter the neighboring instance of $x$ until a counterfactual $x'$ is reached. Although it does not offer a sequence of actions, CERTIFAI generates a sequence of intermediate instances between $x$ and $x'$. CERTIFAI starts by generating a set $I$ of neighboring instances of $x$, which it then iteratively alters through selection, mutation, and crossover until an instance that satisfies the validity constraint is found. Since CERTIFAI forcibly changes feature values, it may generate unlikely counterfactuals that do not conform to causal constraints between the features.

\citet{karimi2021algorithmic} observe that, to generate recourse, causal relationships between features have to be taken into account. The authors propose MINT, a method for finding a minimal set of causal interventions to change the original instance $x$ into a counterfactual $x^{'}$. To find $x^{'}$, the authors use the \textit{Abduction-Action-Prediction} \cite{pearl2018book} procedure, used to generate a counterfactual instance in causal theory. The obtained counterfactual $x'$ obeys the causal constraints in the environment and can be reached from the original instance $x$ using causal interventions. However, MINT requires access to a full causal model of the task, which is often not available.

\citet{karimi2020algorithmic} extend the work of \citet{karimi2021algorithmic} and prove that recourse cannot be guaranteed if the full causal model is not known. \citet{karimi2020algorithmic} propose a probabilistic approach that can generate recourse with high probability, given only the causal graph, without structural equations. To learn an approximate SCM using the training data, a Bayesian approach is proposed, and probabilistic regression was used to learn approximate structural equations.

\begin{table}[t]
    \caption{Evaluation of state-of-the-art counterfactual methods in supervised learning based on whether they address the desired properties.}
    \centering
    \begin{adjustbox}{width=\textwidth}
        \begin{tabular}{@{}lccccccc@{}}
        \toprule
        \textbf{Method} & \textbf{Proximity} & \textbf{Validity} & \textbf{Actionability} & \textbf{Sparsity} & \textbf{Data manifold closeness} & \textbf{Causality} & \textbf{Recourse}\\ \midrule
        
        \citet{wachter2017counterfactual} &  \cmark & \cmark & \xmark & \cmark & \xmark & \xmark & \xmark \\ \midrule
        
        \citet{dandl2020multi} &  \cmark & \cmark & \cmark & \cmark & \cmark & \xmark & \xmark\\ \midrule
        
        \citet{looveren2021interpretable} &  \cmark & \cmark & \cmark & \cmark & \cmark & \xmark & \xmark\\ \midrule
        
        DICE \cite{mothilal2020explaining} &  \cmark & \cmark & \cmark & \cmark & \xmark & \cmark & \xmark\\ \midrule
        
        \citet{laugel2017inverse} &  \cmark & \cmark & \xmark & \cmark & \xmark & \xmark & \xmark\\ \midrule
        
        REVISE \cite{joshi2019towards} &  \cmark & \cmark & \cmark & \xmark & \cmark & \cmark & \cmark\\ \midrule
        
        FACE \cite{poyiadzi2020face} &  \cmark & \cmark & \cmark & \xmark & \cmark & \xmark & \xmark\\ \midrule

        LORE \cite{1283} & \cmark & \cmark & \cmark & \xmark & \cmark & \xmark & \xmark\\ \midrule
        
        ReLACE \cite{chen2021relace} &  \cmark & \cmark & \cmark & \cmark & \xmark & \xmark & \cmark\\  \midrule
        
        \citet{samoilescu2021model} &  \cmark & \cmark & \cmark & \cmark & \cmark &  \xmark & \xmark \\\midrule
        
        \citet{mahajan2019preserving} & \cmark & \cmark & \cmark & \xmark & \cmark & \cmark & \xmark \\ \midrule 
        
        \citet{ustun2019actionable} & \cmark & \cmark & \cmark & \xmark & \xmark & \xmark & \cmark\\ \midrule
        
        \citet{karimi2021algorithmic} & \cmark & \cmark & \cmark & \xmark & \xmark & \cmark & \cmark\\ \midrule
        
        CERTIFAI \cite{sharma2019certifai} & \cmark & \cmark & \cmark & \xmark & \xmark & \xmark & \cmark \\
        
        \bottomrule
        \end{tabular}
        \label{evaluation}
    \end{adjustbox}
\end{table}

\subsection{Evaluation}

Table \ref{evaluation} summarizes how different methods reviewed in this section address counterfactual properties. All methods focus on providing a valid counterfactual and ensuring their proximity to the original instance. However, there is no unified way to evaluate proximity, and different methods chose different metrics to calculate the similarity between the instances, making the methods difficult to compare. Additionally, many methods address the issue of actionability by allowing the user to mark features that should not be changed. Around half of the methods also implement a way of ensuring sparsity, usually by minimizing the number of feature changes. Data manifold closeness has been interpreted in various ways in the current work. While some works simply observe the distance between the counterfactual and the nearest instance \cite{dandl2020multi}, other works measure sample density along the path between the instances \cite{poyiadzi2020face} or represent the space of realistic instances by encoding the training data set \cite{joshi2019towards,samoilescu2021model}. This indicates that there is no common understanding of what data manifold closeness is, and that the notion of a realistic instance remains open to interpretation. 

The least considered properties by far are causality and recourse. Causality is, in a way, a prerequisite for useful recourse generation, because a sequence of actions provided by recourse needs to be causally correct to be useful to the user. Ensuring causality is the most challenging task of generating counterfactuals, as it requires causal information about the environment. However, the causal graph of the environment is rarely available, and often needs to be hand-crafted, requiring human time and effort.

\section{Counterfactual Explanations in Reinforcement Learning}
\label{CFRL}

In the previous section, we provided an overview of the state-of-the-art methods for generating counterfactual explanations in supervised learning tasks. In RL, however, counterfactual explanations have been seldom applied. In this section, we explore how counterfactual explanations differ between supervised and reinforcement learning tasks, and identify the main challenges that prevent the adoption of methods from supervised to reinforcement learning. Additionally, we redefine counterfactual explanations and counterfactual properties for RL. Finally, we identify the main research directions for implementing counterfactual explanations in RL.

\subsection{Existing Methods for Counterfactual Explanations in RL}

Although the potential of counterfactual explanations for explaining RL agents has been recognized \cite{broadXAI}, there are currently few methods enabling counterfactual generation in RL.
\citet{olson2019counterfactual} proposed the first method for generating counterfactuals for RL agents. In their work, counterfactuals are generated using a generative deep-learning architecture. The authors explain the decisions of RL agents in Atari games and search for counterfactuals in the latent space of the policy network. The approach assumes that policy is represented by a deep neural network and starts by dividing the network into two parts. The first part $A$ contains all but the last hidden layer of the network and serves for mapping input state $s$ into latent representation $z$: $A(s) = z$. The second part $\pi$ consists only of the last fully connected layer followed by a softmax and is used to provide action probabilities based on the latent state. The policy network can be imagined as a composition of the network parts $\pi(A(s))$. To generate counterfactual states, authors propose a deep generative model consisting of an encoder ($E$), discriminator ($D$), generator ($G$) and Wasserstein encoder ($E_w, D_w$). Encoder $E$ and generator $G$ work together as an encoder-decoder pair, with the encoder mapping the input state into a low-dimensional latent representation, and the generator performing reconstruction from encoding to the original state. Discriminator $D$ learns to predict the probability distribution over actions $\pi(z)$ given a latent state representation $z = E(s)$. Counterfactuals are searched for in the latent space, which may have holes, resulting in unrealistic instances. For this reason, the authors include a Wasserstein encoder $(E_w, D_w)$ to map latent state $z$ to an even lower-dimensional representation $z_w$ to obtain a more compact format and ensure more likely counterfactuals. The generative architecture is trained on a set of agent's transitions in the environment $\mathcal{X} = \{(s_1, a_1),\dots,(s_n, a_n)\}$. Finally, to generate counterfactual state $s'$ of state $s$, authors first locate the latent representation $z_w^*$ of $s'$ by solving:

\begin{equation}
\begin{split}
     \textrm{minimize }  \quad \quad|&|E_w(A(s)) - z_w^*||_2^2 \\
     \textrm{subject to }\quad \quad &\arg \max_{a \in A} \pi(D_w(z_w^*, a)) = a'
\end{split}
\end{equation}
  
where $a'$ is the counterfactual action. The objective can be simplified to:
\begin{equation}
    \label{counterfactuals-atari}
    z_w^* = \arg \min_{z_w} ||z_w - E_w(A(s))||_2^2 + \log(1 - \pi(D_w(z_w), a'))
\end{equation}

The process for finding the counterfactual instance then consists of encoding the original state into Wasserstein encoding $z_w = E_w(A(s))$, optimizing Equation \ref{counterfactuals-atari} to find $z_w^*$. The latent counterfactual instance $z_w^*$ can then be decoded to obtain $\pi(D_w(z_w^*))$, which is finally passed to the generator to obtain the counterfactual instance $s'$. This approach can generate realistic image-based counterfactual instances that can be used to explain the behavior of Atari agents. However, this approach is model-specific and requires access to RL model's internal parameters. 

In contrast, \citet{huber2023ganterfactual} propose GANterfactual-RL, a model-agnostic generative approach based on the StarGAN \cite{choi2018stargan} architecture for generating realistic counterfactual explanations. \citet{huber2021benchmarking} approach the counterfactual search problem as the domain translation task where each domain is associated with states in which the agent chooses a specific action. A suitable counterfactual can be found by translating the original state to a similar state from the target domain. In this way, obtaining a counterfactual involves only changing the features that are relevant to the agent's decision, while maintaining others. The approach consists of training two neural networks -- discriminator $D$ and generator $G$. The generator $G$ is trained to translate the input state $x$ into $y$, conditioned on the target domain $c$. The role of the discriminator is to distinguish between real states and fake ones produced by $G$.

The current approaches for generating counterfactuals in RL do not deviate from similar approaches in supervised learning and do not account for the additional complexities of RL framework presented in Section \ref{vs}. For example, the counterfactuals generated by \citet{olson2019counterfactual, huber2021benchmarking} rely only on state features to explain a decision and do not include different explanation components, such as goals or objectives that can influence the agent's behavior. Similarly, the approaches do not address the issue of temporality and can generate counterfactuals that are close together in terms of features but might be far away in execution. 

While current methods in supervised and RL  generate counterfactuals suitable for one-step prediction tasks, they do not account for the explanatory requirements of RL tasks. In the following sections, we redefine counterfactual explanations to fit within the RL framework and explore their desired properties. We also identify the main research directions that need to be addressed before counterfactuals can be successfully used to explain the decisions of RL agents.

\subsection{Counterfactual Explanations Redefined}

In RL, current counterfactual explanations focus on interpreting a single decision in a specific state \cite{olson2019counterfactual,huber2023ganterfactual}. Variables are simply state features, and the score is an agent's choice of action in the state. This corresponds to the definition of counterfactuals used in supervised learning, where input features are replaced by state features and a model's prediction of a class label with an action prediction. However, unlike supervised learning models, RL agents deal with sequential tasks where states are temporally connected. This means that an agent's action in a single state cannot be fully explained by taking into account only the current state features. Additionally, an agent's behavior cannot be fully explained by interpreting single actions, as individual actions are often part of a larger plan or policy. To be able to generate counterfactual explanations that can fully encompass the complexities of RL tasks, we need to redefine the notions of variable and score from the RL perspective. Examples of possible counterfactual explanations in RL after redefining the terms variable and score are presented in Table \ref{cf_explanations}.

\begin{table}[t]
    \caption{Different types of counterfactual explanations in RL depending on the definition of variable and score.}
    \centering
    \begin{adjustbox}{width=\textwidth}
        \begin{tabular}{@{}lccc@{}}
        \toprule
        & \textbf{Action} & \textbf{Plan} & \textbf{Policy} \\ \midrule
        
        Features & \makecell{Had state features taken \underline{different values},\\ \underline{action} $a'$ would be chosen instead of $a$} & \makecell{Had state features taken \underline{different values},\\ \underline{path} $p'$ would be chosen instead of $p$} & \makecell{Had state features taken \underline{different values},\\ \underline{policy} $\pi'$ would be chosen instead of $\pi$}\\ \midrule
        
        Goal & \makecell{Had the agent followed a \underline{different goal} in state $s$,\\ \underline{action} $a'$ would be chosen instead of $a$} &\makecell{Had agent followed a \underline{different goal} in state $s$,\\ \underline{plan} $p'$ would be chosen instead of $p$} & \makecell{Had agent followed a \underline{different goal} in state $s$,\\ \underline{policy} $\pi'$ would be chosen instead of $\pi$}\\ \midrule
        
        Objective & \makecell{Had the agent preferred a \underline{different objective} in state $s$,\\ \underline{action} $a'$ would be chosen instead of $a$} & \makecell{Had agent preferred a \underline{different objective} in state $s$,\\ \underline{plan} $p'$ would be chosen instead of $p$} & \makecell{Had agent preferred a \underline{different objective} in state $s$,\\ \underline{policy} $\pi'$ would be chosen instead of $\pi$} \\ \midrule
        
        Event &  \makecell{Had the agent encountered a \underline{different event} in state $s$,\\ \underline{action} $a'$ would be chosen instead of $a$} & \makecell{Had agent encountered a \underline{different event} in state $s$,\\ \underline{plan} $p'$ would be chosen instead of $p$} & \makecell{Had agent encountered a \underline{different event} in state $s$,\\ \underline{policy} $\pi'$ would be chosen instead of $\pi$}\\
        \midrule
        
        Expectation &  \makecell{Had the agent faced \underline{different expectation} in state $s$,\\ \underline{action} $a'$ would be chosen instead of $a$} & \makecell{Had agent faced \underline{different expectation} in state $s$,\\ \underline{plan} $p'$ would be chosen instead of $p$} & \makecell{Had agent faced \underline{different expectation} in state $s$,\\ \underline{policy} $\pi'$ would be chosen instead of $\pi$}\\
        \bottomrule
        \end{tabular}
    \end{adjustbox}
    \label{cf_explanations}
\end{table}

\subsubsection{Variables}

To redefine variables from the perspective of RL tasks, we explore different possible causes of decisions for RL agents. According to the causal attribution model presented by \citet{broadXAI}, agent's decisions can be directly or indirectly affected by $5$ factors -- \textit{perception, goals, disposition, events and expectations}. While \citet{broadXAI} also provides a causal structure connecting different factors, in this work we focus only on their individual effect on the agent's decision.  

Using an agent's perception as a variable corresponds to the current work in counterfactual explanations focuses on state features as the sole causes of a decision in a state. Counterfactual question can then be phrased as \textit{``Given that agent chose action a in state with state features ${f_1, ..., f_k}$ how can the features be changed for the agent to choose an alternative action a'?''} This type of explanation is most suitable for single-goal, single-objective, deterministic tasks, where each action depends only on the current state agent is perceiving. In such tasks, it can be assumed that user knows the ultimate goal of the agent's behavior and the agent does not need to balance between different objectives. In such environments, an agent's actions have deterministic consequences and unexpected events cannot influence the agent's behavior. This corresponds to simple environments, such as deterministic games, where the agent's goal is to win by optimizing a single objective function. In more complex, real-world environments, however, relying only on the current state to make a decision is unlikely, and additional factors might need to be taken into account. 

An agent's current goal can also influence an outcome. Counterfactual explanations can then be used to address the question \textit{``Given that agent chose action a in a state s while following goal G, for what alternative goal G' would agent have chosen action a'?''}. This question is especially useful in multi-goal or hierarchical RL tasks, where the agent has a set of goals that need to be fulfilled and can switch between them. In such environments, knowing which goal an agent is pursuing in a state is necessary to fully understand its behavior. For example, consider an autonomous taxi vehicle. The task of the taxi agent can be split into two subgoals -- picking up passengers and delivering them to their destinations. A passenger picked up in such a taxi might be confused if the agent takes a longer route to their destination. The passenger might lose trust in the agent and even suspect the agent is lost or is trying to trick the user. However, if an agent is allowed to explain its behavior by employing a counterfactual statement: \textit{``I am following a goal of picking up a different passenger. Had I followed a goal of delivering the current passenger I would have taken the shorter route to the destination.''}, the user can be reassured that the agent is indeed following the best course of action.

Similarly, in multi-objective environments, an agent's preference for a specific objective can guide its behavior. This corresponds to the effect of the agent's disposition on its decisions as described in \citet{broadXAI}. The counterfactual question can then be posed as: \textit{``Given that agent chose action a in state s while preferring objective O, for what alternative objective preference O' would agent have chosen action a'?''}. An agent's internal preference for one objective over another is likely to influence its actions, and this information is necessary to fully understand the agent's behavior. Many real-life tasks fit this description. For example, the behavior of an autonomous driving agent is at each time step guided by different objectives such as speed, safety, fuel consumption, user comfort... Different agents can have different preferences in regard to these objectives. If an agent does not slow down in a critical situation, and as a consequence ends up colliding with the vehicle in front of it, the counterfactual explanation could identify that, had the agent preferred safety over speed in the critical state, it would have chosen to slow down and the accident would have been avoided. This type of explanation can also be useful for developers when designing multi-objective reward functions for complex tasks. Deciding on the correct weights for different objectives is a notoriously challenging task and requires substantial time and effort \cite{liu2014multiobjective}. Understanding how different preferences of objectives influence an agent's decision, especially in critical states, can be useful for adjusting the appropriate weights. 

Unexpected outside events can alter an agent's behavior and as such should also be featured in counterfactual explanations. The counterfactual question can then be defined as \textit{``Given that agent chose action a in state under an unexpected event E, for which event E' would agent have chosen action a'?''}. For example, consider an autonomous driving agent that had to change its trajectory due to an unexpected blockage on the road. The user might be confused as to why the agent decided on the longer path, and counterfactual explanations could restore their trust in the system by asserting that \textit{``Had the shorter path not been blocked, I would have taken it.'' }. This corresponds to the field of safe RL, which argues that, to be safely deployed, agents must be able to avoid danger and correctly respond to unexpected events \cite{garcia2015comprehensive}. 

Finally, expectations can also influence an agent's behavior. Expectation refers to a set of social and cultural constraints that are imposed on the agent and can affect its decisions \cite{broadXAI}. These can be anything from social conventions and laws to ethical rules. Counterfactual explanations could focus on how changing expectations affects an agent's behavior by posing the question \textit{``Had expectations for the agent been different, would it still make the same decisions?}''. Including expectations in the explanations can help frame an agent's behavior in the broader social and cultural context. Users who are not familiar with the expectations the agent is implementing may find it strange if the agent diverges from its primary goals due to an expectation. Explanations that identify social expectations as the cause of unexpected behavior can help users understand and regain trust in the model.  

While decisions of a supervised learning model depend only on the input features, the RL agent's behavior is often motivated by a wider range of causes, such as goals, objectives, and expectations. For that reason, the definition of a variable needs to be extended to include all factors influencing an agent's decisions. 

\subsubsection{Score}
In supervised learning, the only reasonable outcome of the model's behavior for a specific input instance is its prediction. In RL tasks, however, there are multiple possible scores that could describe an agent's behavior in a specific state. Current methods for counterfactual explanations in RL use the action chosen in the state as the score. However, each agent's decision often does not exist on its own, but as a part of a larger plan or a policy. Relying only on a current action might not be enough to capture the desired change in behavior, especially in complex, continuous environments. In game environments, such as chess or StarCraft, the user might be more interested to know why an agent chose one specific plan or sequence of steps instead of another. Similarly, in an autonomous driving task with continuous control, the user is unlikely to ask the agent why it changed the steering angle by a small amount but might be interested to know why it chose a specific route over another. Interpreting a single action in a complex or continuous environment might be too low-level for non-expert users, and better suited for experts or developers. Non-expert users, however, tend to be more comfortable with high-level explanations, corresponding to plans or policies. Counterfactual explanations in RL have to be extended to include more complex behavior that is inherent to RL agents to be user-friendly.

While supervised learning models focus on one-step prediction tasks, in RL agents learn sequential behavior which consists of plans and policies. For this reason, counterfactual explanations need to account for the broader scope of RL behavior. 

\subsection{Counterfactual Properties Redefined}

Counterfactual properties are metrics for deciding on the most suitable counterfactual to be presented to the user. It is of high importance that counterfactual properties describe necessary desiderata for choosing useful counterfactuals. Providing a user with unattainable or ineffective counterfactuals can cause frustration and deteriorate user trust in the system. So far, counterfactual properties have been defined with supervised learning tasks in mind, where only input features are changed in order to obtain a desired prediction. Since counterfactual explanations in RL are not limited only to state features and one-step decisions, counterfactual properties need to be redefined for use in RL tasks:

\begin{enumerate}
    \item \textit{Validity} is measured as the extent to which the desired score is achieved in the counterfactual instance. In RL, the score can refer to the agent's current action, plan, or policy. Depending on the chosen definition of the score, validity describes whether the agent in the counterfactual instance chose the desired action or followed the desired plan or policy. 
    
    \item \textit{Proximity} is estimated as a difference between the original and counterfactual instances. In supervised learning, similarity is calculated as a function of input features. In RL however, proximity can include not only state features but other explanation factors, such as goals or objectives, which also need to be included when calculating proximity. Additionally, while in supervised learning data samples are independent of each other, states in RL tasks are temporally connected. Offering the user a counterfactual instance that is similar to the original in terms of state features, but temporally distant might be useless. For this reason, temporality also must be considered when calculating the proximity of two states in RL.
    
    \item \textit{Actionability} specifies the need for allowing certain features (e.g. race, country of origin) to be immutable. In supervised learning, developers are required to define which features should not be changed. In RL however, certain immutable features can also be contained within the environment itself. For example, although goal coordinates exist in the feature space, changing them might be impossible under the rules of the environment. Suggesting to the user that they should obtain a counterfactual where the goal is moved would not be actionable. For this reason, the definition of actionability in RL needs to be expanded to limit change of not only developer-defined features but also those that are rendered immutable by the rules of the environment.
    
    \item \textit{Sparsity} refers to the fact that as few as possible features should be changed in order to obtain the counterfactual instance. In current work, it is often assumed that there is no difference in the cost of changing individual features. In RL, however, counterfactual instances can be obtained not only by changing state features but also goals or objectives. Changing a single state's features or changing the agent's goal may pose different levels of difficulty for the user, and a sparsity metric should reflect this potential cost inequality in RL tasks. 
    
    \item \textit{Data manifold closeness} refers to the requirement that counterfactual instances should be realistic, in order for users to be able to achieve them. The main challenge for estimating data manifold closeness is in defining what is meant by the notion of a realistic instance. In supervised learning, the training data set is used to represent the space of realistic instances. The data manifold closeness of an instance can then be estimated with respect to the defined realistic sample space. In RL, however, the training data set is not readily available. For this reason, a more precise definition of realistic instances in RL and accompanying metrics for evaluating data manifold closeness are needed in RL. 
    
    \item \textit{Causality} refers to the notion that counterfactual instances should obey causal constraints between features. In supervised learning, generating causally correct counterfactual instances is only possible if the causal relationships between input features are known. In RL, however, causal rules are ingrained in the environment. Relying on actions provided by the environment can ensure that features are not changed independently but in accordance with the causal rules of the environment. 
    
    \item \textit{Recourse} is a sequence of actions the user should perform to transform the original instance into a counterfactual. In supervised learning, the user is in theory allowed to perform whichever actions they desire, provided that they conform to the causality and actionability constraints. In certain RL tasks, however, the user's actions might be limited to those available in the environment. In game environments, for example, the user can only change the game state by performing actions that obey the rules of the game. For illustration, consider the game StarCraft, a two-player strategy game where the goal is to build an army and defeat the opponent. If the user is interested to know why RL agent did not choose an \textit{attack} action in a specific state, the system can answer by providing recourse: \textit{``In order for \textit{attack} to be the best decision, you first need to perform action \textit{build\_army}.''} To follow the provided advice, the user must rely only on actions available in the environment.
\end{enumerate}

Counterfactual properties guide the search for the counterfactual instance, and as such directly influence which explanations will be presented to the user. In RL, counterfactual properties need to be extended from their supervised learning counterparts, to encapsulate the complexities of RL tasks. The introduction of different types of variables and scores to counterfactual explanations in RL affects all counterfactual properties. Additionally, the temporal nature of RL needs to be considered when evaluating proximity, one of the properties that all previous works optimize. On the other hand, while supervised learning lacks causal information about input features, this knowledge is often ingrained within an RL environment, and can potentially be used to generate causally-correct counterfactuals and recourse.

\subsection{Challenges and Open Questions}

So far, we have shown how counterfactual explanations differ between supervised and reinforcement learning and have redefined them for RL tasks. However, the question remains, how can suitable counterfactuals be generated for RL tasks? As there is a large amount of research on counterfactual explanations in supervised learning, and due to similarities between the two learning paradigms, in this section we explore whether this substantial research can be applied to RL. We identify the main challenges and open research directions on this path that should be tackled before counterfactual explanations can be generated for RL tasks.

\subsubsection{Counterfactual Search Space}

In supervised learning, the search for a counterfactual instance is often performed over the training data set. The training data set is an essential component of every supervised learning task. Additionally, the training data set is used to represent the space of realistic instances, to measure the data manifold closeness of counterfactual instances. Intuitively, if an instance is consistent with the data distribution defined by the training data set, it is considered realistic. In RL, however, the agent learns the task from scratch and the training data set is not available. For that reason, the search space corresponding to realistic instances needs to be redefined for RL tasks. 

The most straightforward way to approximate training data set in RL is to obtain a data set of states by unrolling the agent's policy. Existing loss-based approaches can then be optimized over such a data set to find the most useful counterfactual instance. Data manifold closeness of counterfactual instances could then be estimated in relation to this data set and realistic instances would correspond to those states that the agent is likely to visit by following its policy. This approach makes sense for counterfactual instances which can be obtained by only changing the state features. The counterfactual instance should then be reachable by the same policy the agent is employing. However, if searching for it includes changing the agent's goals or dispositions, the counterfactual instance is not likely to be found within the agent's current policy. For example, consider a multi-goal agent with two subgoals $A$ and $B$, explaining why decision $a'$ was not taken instead of $a$ in state $s$. A counterfactual instance might explain the decision by claiming that, had the agent been following subgoal $B$ instead of $A$, action $a'$ would have been executed instead of action $a$. However, this counterfactual instance is unlikely to occur under the agent's policy $\pi$ which is already following goal $A$ in state $s$. For this reason, further research to understand how can the search space for counterfactual instances be defined is needed.

\subsubsection{Categorical Variables}

Most of the current state-of-the-art methods for generating counterfactual explanations in supervised learning do not support categorical input features. This problem naturally extends to RL as well, where state features can often be categorical. However, in RL this problem is additionally exacerbated with the use of alternative explanation components such as goals, dispositions, or events. These components often represent high-level concepts and as such are often discrete. Understanding how discrete variables can be integrated with existing counterfactual methods is necessary for their employment in RL tasks.

\subsubsection{Temporality}
One of the main purposes of counterfactual explanations is to provide a user with a way to change the input features in the most suitable way to change the output of a black-box model. Intuitively, this means that the counterfactual needs to be reachable from the original instance. In supervised learning, many counterfactual properties such as proximity, sparsity, and data manifold closeness are used to guide the search toward easily obtainable counterfactuals. In supervised learning, the notion of reachability is not contained within the task -- all samples are considered independent, and whether one sample can be obtained from another often depends on human-defined interpretation of terms such as immutability, causality, or sparsity. In most supervised learning works, proximity is used as a proxy for reachability, and states with similar state features are considered easily reachable. In RL, however, due to its sequential nature, states are temporally connected. Reachability in RL tasks is ingrained in the environment and depends on how temporally distant two states are in terms of execution. 

\begin{figure}[t]
    \centering
    \includegraphics[width=0.3\textwidth]{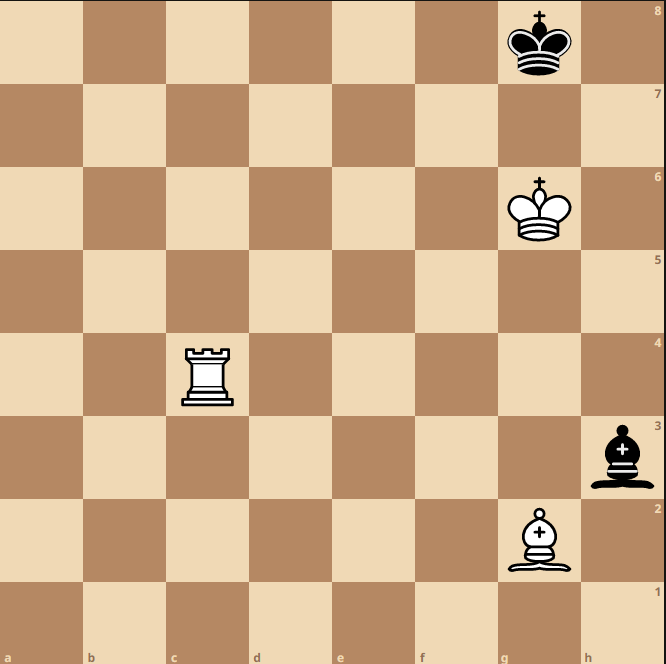}
    \quad \quad \quad
    \includegraphics[width=0.3\textwidth]{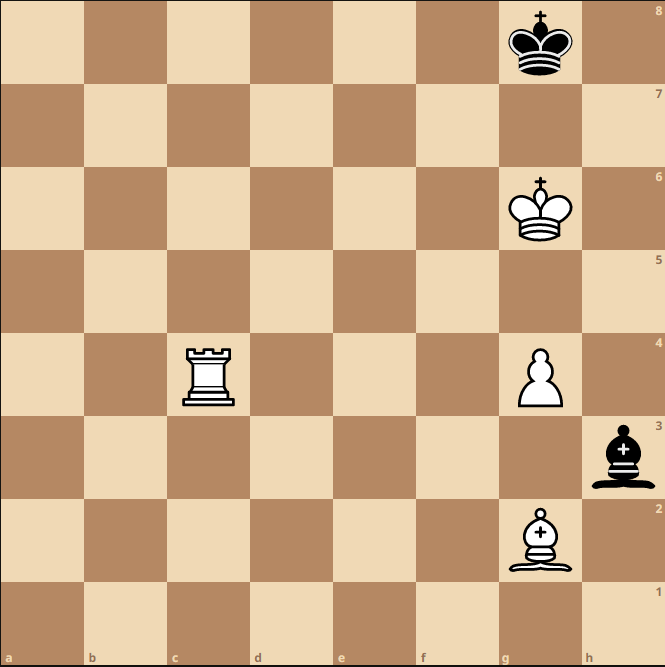}
    \caption{\textbf{Left}: original game state $x$. \textbf{Right}: cf game state $x'$ obtained through FACE algorithm \cite{poyiadzi2020face}, answering the question \textit{``In what position would c4c8 be the best move?''}. While the provided counterfactual is similar to the original state based on its state features, it is not reachable from the original state using the rules of the game, and as such does not offer actionable advice to the user.}
    \label{temporality}
\end{figure}

This means that two states with almost identical state feature values might be distant from each other in terms of execution. For illustration, consider an example of a chess game presented in Figure \ref{temporality}. The generated counterfactual for a chess position can satisfy validity, sparsity, and data manifold closeness constraints and be very similar to the original instance in terms of state features. However, if it is not reachable from the original instance using the game rules, recourse cannot be provided based on such counterfactual. In other words, the counterfactual is not useful to the user if they cannot reach it by following the game rules. For this reason, temporal similarity needs to be considered along with the similarity of features when calculating proximity. Further research is necessary to define metrics and integrate temporal similarity in the search for counterfactuals in RL.

\subsubsection{Stochasticity}

Most methods for providing recourse in supervised learning assume that model decisions and feature changes are deterministic and that there exists one true sequence of actions for obtaining the counterfactual instance. In many RL tasks, however, the environment can exhibit stochastic behavior and influence the agent's behavior. Events outside of the agent's control can change the environment, and the agent's actions can have stochastic consequences. This is also a characteristic of real-life problems, where decision models can change through time, and performing an action might not have a deterministic outcome. For example, the threshold for a loan can change over time, and even if the user follows the recourse offered a year ago, their application might still get rejected. This inconsistency between advice and results can be frustrating for the user and undermine their trust in the system. For that reason, counterfactual explanations in RL should account for the stochastic nature of the environment when offering recourse to users. For example, counterfactual search could incorporate optimization of certainty constraint, ensuring that the user is presented not only with the fastest but also the most secure path towards their desired outcome.

\subsubsection{Alternative Explanation Components}

Current methods for counterfactual explanations in RL rely only on state features as causes of a decision. However, in RL, various other components such as goals, objectives, or outside events can contribute to the agent's behavior and should be captured in counterfactual explanations. Especially in the fields of multi-goal and multi-objective RL, diverse counterfactual explanations that capture the wide range of possible causes are necessary in order to fully comprehend the agent's decisions. Further research in the integration of different causes into counterfactual explanations is needed.

Developing metrics for evaluating counterfactual properties in RL is another important research direction. So far, counterfactual properties have been defined and evaluated with supervised learning in mind. However, research is required into how redefined counterfactual properties for RL tasks can be evaluated. For example, to calculate proximity for RL states, temporal similarity might need to be considered. However, it is not clear how to efficiently estimate how easily a state is reachable from another state in RL. A brute-force approach would include a nearest-path search over the graph of all possible transitions in the environment. However, in high-dimensional environments, this approach would be prohibitively expensive, and distances between states would need to be estimated. Similarly, research is needed in estimating data manifold closeness for counterfactual instances in RL. While a training data set can be used in supervised learning to represent the space of realistic instances, in RL tasks there is no natural equivalent to the training data set. For that reason, an alternative metric for evaluating data manifold closeness is necessary. For purposes of evaluating sparsity and proximity, developing metrics that integrate state features with RL-specific explanations components like goals and objectives, is another research direction necessary for implementing counterfactual explanations in RL. Existing research in reward function similarity \cite{gleave2020quantifying} could be the starting point for the development of metrics for comparing the goals and objectives of RL agents. Additionally, the cost associated with changing individual goals and objectives when generating recourse might depend on the user. This opens the research to the possibility of a human-in-the-loop approach \cite{arzate2020survey} where user feedback could be integrated into the counterfactual generation process.   

\subsubsection{Counterfactual and Prefactual Statements}

Previous research in psychology distinguishes between prefactual and counterfactual explanations, two different types of statements that can be used to describe causes of an event \cite{dai2022counterfactual,ferrante2013improving,byrne2004counterfactual}. For example, for a user wondering why their loan application has been rejected, a counterfactual explanation could state: \textit{Had you had a higher income, your loan would have been approved}, while the prefactual explanation would advise: \textit{If you obtain a higher income in the future, your next loan application will be approved} \cite{dai2022counterfactual}. Previous research in psychology has found that the two explanation types include different psychological processes \cite{ferrante2013improving,byrne2004counterfactual}.

In supervised learning, where models deal with one-step predictions, and there is no notion of time, counterfactual and prefactual statements are used interchangeably. In contrast, RL is designed for sequential tasks, where the distinction between prefactual and counterfactual statements makes more sense.
For example, while counterfactual explanations could help locate mistakes in the agent's previous path, prefactual explanations could help guide them from an undesirable state to a more favorable one. In this sense, counterfactual explanations could be related to the field of credit assignment \cite{harutyunyan2019hindsight}. Prefactual explanations, on the other hand, could be used in the field of safe RL to help agents avoid dangerous states and find a way to safety \cite{garcia2015comprehensive}.

While the difference between prefactual and counterfactual statements in supervised learning is purely semantic, in RL different technical approaches would need to be developed to obtain the two types of explanations. Current methods for counterfactual generation in RL \cite {olson2019counterfactual,huber2023ganterfactual} do not distinguish between prefactual and counterfactual statements and further research is needed to utilize these two perspectives of counterfactual explanations.

\subsubsection{Evaluation}
Evaluating explanations is challenging as their perceived quality is subjective and depends on the user. Counterfactual explanations are usually evaluated based on their counterfactual properties. Few works build on this and evaluate the real-life applicability of counterfactuals in a user study \cite{olson2019counterfactual}. In supervised learning, counterfactual explanations are evaluated against data sets such as German credit \cite{dandl2020multi,mothilal2020explaining}, Breast Cancer \cite{chen2021relace,looveren2021interpretable} or MNIST \cite{sharma2019certifai,looveren2021interpretable,samoilescu2021model,laugel2017inverse}. For RL tasks, however, there is no established benchmark environment for the evaluation of counterfactual explanations. Current work in RL evaluates counterfactuals in Atari games \cite{olson2019counterfactual}.  Establishing benchmark environments is necessary for evaluating and comparing counterfactual methods in RL. Additionally, in RL benchmark environments need to be extended to include multi-goal, multi-objective, and stochastic tasks.

\subsection{Conclusion}
While successfully applied to a variety of tasks, RL algorithms suffer from a lack of transparency due to their reliance on neural networks. User-friendly explanations are necessary to ensure trust and encourage collaboration of non-expert users with the black-box system. In this work, we explored counterfactuals -- a user-friendly, actionable explanation for interpreting black-box systems. Counterfactuals represent a powerful explanation method that can help users understand and better collaborate with the black-box system. However, counterfactuals that offer unrealistic changes or do not deliver a desired output can further damage user trust and hinder their engagement with the system. For this reason, only well-defined and useful counterfactual explanations must be presented to the user. 

Although they are researched in supervised learning, counterfactual explanations are still underrepresented in RL tasks. 
In this work, we explored how counterfactual explanations can be redefined for RL tasks. Firstly, we offered an overview of current state-of-the-art methods for generating counterfactual explanations in supervised learning. Furthermore, we identified the main differences between supervised and RL from the perspective of counterfactual explanations and provided a definition more suited for RL tasks. Specifically, we recognized that definitions of score and variable are not straightforward in RL, and can encompass different concepts. Finally, we proposed the main research directions that are necessary for the successful implementation of counterfactual explanations in RL. Specifically, we identified temporality and stochasticity as important RL-specific concepts that affect counterfactual generation. Additionally, we brought attention to universal issues with counterfactual generation both in supervised and RL, such as the handling of categorical features and evaluation. 

\section*{Acknowledgement}
This publication has emanated from research conducted with the financial support of a grant from Science Foundation Ireland under Grant number 18/CRT/6223 and SFI Frontiers for the Future grant number 21/FFP-A/8957. For the purpose of Open Access, the author has applied a CC BY public copyright license to any Author Accepted Manuscript version arising from this submission. We thank James McCarthy for helpful feedback on the early versions of the manuscript.

\bibliographystyle{ACM-Reference-Format}
\bibliography{sample-base}


\begin{thebibliography}{102}


\ifx \showCODEN    \undefined \def \showCODEN     #1{\unskip}     \fi
\ifx \showDOI      \undefined \def \showDOI       #1{#1}\fi
\ifx \showISBNx    \undefined \def \showISBNx     #1{\unskip}     \fi
\ifx \showISBNxiii \undefined \def \showISBNxiii  #1{\unskip}     \fi
\ifx \showISSN     \undefined \def \showISSN      #1{\unskip}     \fi
\ifx \showLCCN     \undefined \def \showLCCN      #1{\unskip}     \fi
\ifx \shownote     \undefined \def \shownote      #1{#1}          \fi
\ifx \showarticletitle \undefined \def \showarticletitle #1{#1}   \fi
\ifx \showURL      \undefined \def \showURL       {\relax}        \fi
\providecommand\bibfield[2]{#2}
\providecommand\bibinfo[2]{#2}
\providecommand\natexlab[1]{#1}
\providecommand\showeprint[2][]{arXiv:#2}

\bibitem[Abiodun et~al\mbox{.}(2018)]%
        {abiodun2018state}
\bibfield{author}{\bibinfo{person}{Oludare~Isaac Abiodun},
  \bibinfo{person}{Aman Jantan}, \bibinfo{person}{Abiodun~Esther Omolara},
  \bibinfo{person}{Kemi~Victoria Dada}, \bibinfo{person}{Nachaat~AbdElatif
  Mohamed}, {and} \bibinfo{person}{Humaira Arshad}.}
  \bibinfo{year}{2018}\natexlab{}.
\newblock \showarticletitle{State-of-the-art in artificial neural network
  applications: A survey}.
\newblock \bibinfo{journal}{\emph{Heliyon}} \bibinfo{volume}{4},
  \bibinfo{number}{11} (\bibinfo{year}{2018}), \bibinfo{pages}{e00938}.
\newblock


\bibitem[Amir and Amir(2018)]%
        {amir2018highlights}
\bibfield{author}{\bibinfo{person}{Dan Amir} {and} \bibinfo{person}{Ofra
  Amir}.} \bibinfo{year}{2018}\natexlab{}.
\newblock \showarticletitle{Highlights: Summarizing agent behavior to people}.
  In \bibinfo{booktitle}{\emph{Proceedings of the 17th International Conference
  on Autonomous Agents and MultiAgent Systems}}. \bibinfo{pages}{1168--1176}.
\newblock


\bibitem[Amodei et~al\mbox{.}(2016)]%
        {amodei2016concrete}
\bibfield{author}{\bibinfo{person}{Dario Amodei}, \bibinfo{person}{Chris Olah},
  \bibinfo{person}{Jacob Steinhardt}, \bibinfo{person}{Paul Christiano},
  \bibinfo{person}{John Schulman}, {and} \bibinfo{person}{Dan Man{\'e}}.}
  \bibinfo{year}{2016}\natexlab{}.
\newblock \showarticletitle{Concrete problems in AI safety}.
\newblock \bibinfo{journal}{\emph{arXiv preprint arXiv:1606.06565}}
  (\bibinfo{year}{2016}).
\newblock


\bibitem[Aradi(2020)]%
        {aradi2020survey}
\bibfield{author}{\bibinfo{person}{Szil{\'a}rd Aradi}.}
  \bibinfo{year}{2020}\natexlab{}.
\newblock \showarticletitle{Survey of deep reinforcement learning for motion
  planning of autonomous vehicles}.
\newblock \bibinfo{journal}{\emph{IEEE Transactions on Intelligent
  Transportation Systems}} \bibinfo{volume}{23}, \bibinfo{number}{2}
  (\bibinfo{year}{2020}), \bibinfo{pages}{740--759}.
\newblock


\bibitem[Arulkumaran et~al\mbox{.}(2017)]%
        {8103164}
\bibfield{author}{\bibinfo{person}{Kai Arulkumaran},
  \bibinfo{person}{Marc~Peter Deisenroth}, \bibinfo{person}{Miles Brundage},
  {and} \bibinfo{person}{Anil~Anthony Bharath}.}
  \bibinfo{year}{2017}\natexlab{}.
\newblock \showarticletitle{Deep Reinforcement Learning: A Brief Survey}.
\newblock \bibinfo{journal}{\emph{IEEE Signal Processing Magazine}}
  \bibinfo{volume}{34}, \bibinfo{number}{6} (\bibinfo{year}{2017}),
  \bibinfo{pages}{26--38}.
\newblock
\urldef\tempurl%
\url{https://doi.org/10.1109/MSP.2017.2743240}
\showDOI{\tempurl}


\bibitem[Arzate~Cruz and Igarashi(2020)]%
        {arzate2020survey}
\bibfield{author}{\bibinfo{person}{Christian Arzate~Cruz} {and}
  \bibinfo{person}{Takeo Igarashi}.} \bibinfo{year}{2020}\natexlab{}.
\newblock \showarticletitle{A survey on interactive reinforcement learning:
  Design principles and open challenges}. In
  \bibinfo{booktitle}{\emph{Proceedings of the 2020 ACM designing interactive
  systems conference}}. \bibinfo{pages}{1195--1209}.
\newblock


\bibitem[Beck and Teboulle(2009)]%
        {beck2009fast}
\bibfield{author}{\bibinfo{person}{Amir Beck} {and} \bibinfo{person}{Marc
  Teboulle}.} \bibinfo{year}{2009}\natexlab{}.
\newblock \showarticletitle{A fast iterative shrinkage-thresholding algorithm
  for linear inverse problems}.
\newblock \bibinfo{journal}{\emph{SIAM journal on imaging sciences}}
  \bibinfo{volume}{2}, \bibinfo{number}{1} (\bibinfo{year}{2009}),
  \bibinfo{pages}{183--202}.
\newblock


\bibitem[Beery et~al\mbox{.}(2018)]%
        {beery2018recognition}
\bibfield{author}{\bibinfo{person}{Sara Beery}, \bibinfo{person}{Grant
  Van~Horn}, {and} \bibinfo{person}{Pietro Perona}.}
  \bibinfo{year}{2018}\natexlab{}.
\newblock \showarticletitle{Recognition in terra incognita}. In
  \bibinfo{booktitle}{\emph{Proceedings of the European conference on computer
  vision (ECCV)}}. \bibinfo{pages}{456--473}.
\newblock


\bibitem[Beyret et~al\mbox{.}(2019)]%
        {beyret2019dot}
\bibfield{author}{\bibinfo{person}{Benjamin Beyret}, \bibinfo{person}{Ali
  Shafti}, {and} \bibinfo{person}{A~Aldo Faisal}.}
  \bibinfo{year}{2019}\natexlab{}.
\newblock \showarticletitle{Dot-to-dot: Explainable hierarchical reinforcement
  learning for robotic manipulation}. In \bibinfo{booktitle}{\emph{2019
  IEEE/RSJ International Conference on Intelligent Robots and Systems (IROS)}}.
  IEEE, \bibinfo{pages}{5014--5019}.
\newblock


\bibitem[Burkart and Huber(2021)]%
        {burkart2021survey}
\bibfield{author}{\bibinfo{person}{Nadia Burkart} {and}
  \bibinfo{person}{Marco~F Huber}.} \bibinfo{year}{2021}\natexlab{}.
\newblock \showarticletitle{A survey on the explainability of supervised
  machine learning}.
\newblock \bibinfo{journal}{\emph{Journal of Artificial Intelligence Research}}
   \bibinfo{volume}{70} (\bibinfo{year}{2021}), \bibinfo{pages}{245--317}.
\newblock


\bibitem[Byrne(2019)]%
        {byrne2019counterfactuals}
\bibfield{author}{\bibinfo{person}{Ruth~MJ Byrne}.}
  \bibinfo{year}{2019}\natexlab{}.
\newblock \showarticletitle{Counterfactuals in Explainable Artificial
  Intelligence (XAI): Evidence from Human Reasoning.}. In
  \bibinfo{booktitle}{\emph{IJCAI}}. \bibinfo{pages}{6276--6282}.
\newblock


\bibitem[Byrne and Egan(2004)]%
        {byrne2004counterfactual}
\bibfield{author}{\bibinfo{person}{Ruth~MJ Byrne} {and}
  \bibinfo{person}{Suzanne~M Egan}.} \bibinfo{year}{2004}\natexlab{}.
\newblock \showarticletitle{Counterfactual and prefactual conditionals.}
\newblock \bibinfo{journal}{\emph{Canadian Journal of Experimental
  Psychology/Revue canadienne de psychologie exp{\'e}rimentale}}
  \bibinfo{volume}{58}, \bibinfo{number}{2} (\bibinfo{year}{2004}),
  \bibinfo{pages}{113}.
\newblock


\bibitem[Cao et~al\mbox{.}(2018)]%
        {cao2018deep}
\bibfield{author}{\bibinfo{person}{Chensi Cao}, \bibinfo{person}{Feng Liu},
  \bibinfo{person}{Hai Tan}, \bibinfo{person}{Deshou Song},
  \bibinfo{person}{Wenjie Shu}, \bibinfo{person}{Weizhong Li},
  \bibinfo{person}{Yiming Zhou}, \bibinfo{person}{Xiaochen Bo}, {and}
  \bibinfo{person}{Zhi Xie}.} \bibinfo{year}{2018}\natexlab{}.
\newblock \showarticletitle{Deep learning and its applications in biomedicine}.
\newblock \bibinfo{journal}{\emph{Genomics, proteomics \& bioinformatics}}
  \bibinfo{volume}{16}, \bibinfo{number}{1} (\bibinfo{year}{2018}),
  \bibinfo{pages}{17--32}.
\newblock


\bibitem[Chen et~al\mbox{.}(2021)]%
        {chen2021relace}
\bibfield{author}{\bibinfo{person}{Ziheng Chen}, \bibinfo{person}{Fabrizio
  Silvestri}, \bibinfo{person}{Gabriele Tolomei}, \bibinfo{person}{He Zhu},
  \bibinfo{person}{Jia Wang}, {and} \bibinfo{person}{Hongshik Ahn}.}
  \bibinfo{year}{2021}\natexlab{}.
\newblock \showarticletitle{ReLACE: Reinforcement Learning Agent for
  Counterfactual Explanations of Arbitrary Predictive Models}.
\newblock \bibinfo{journal}{\emph{arXiv preprint arXiv:2110.11960}}
  (\bibinfo{year}{2021}).
\newblock


\bibitem[Choi et~al\mbox{.}(2018)]%
        {choi2018stargan}
\bibfield{author}{\bibinfo{person}{Yunjey Choi}, \bibinfo{person}{Minje Choi},
  \bibinfo{person}{Munyoung Kim}, \bibinfo{person}{Jung-Woo Ha},
  \bibinfo{person}{Sunghun Kim}, {and} \bibinfo{person}{Jaegul Choo}.}
  \bibinfo{year}{2018}\natexlab{}.
\newblock \showarticletitle{Stargan: Unified generative adversarial networks
  for multi-domain image-to-image translation}. In
  \bibinfo{booktitle}{\emph{Proceedings of the IEEE conference on computer
  vision and pattern recognition}}. \bibinfo{pages}{8789--8797}.
\newblock


\bibitem[Coppens et~al\mbox{.}(2019)]%
        {coppens2019distilling}
\bibfield{author}{\bibinfo{person}{Youri Coppens}, \bibinfo{person}{Kyriakos
  Efthymiadis}, \bibinfo{person}{Tom Lenaerts}, \bibinfo{person}{Ann Now{\'e}},
  \bibinfo{person}{Tim Miller}, \bibinfo{person}{Rosina Weber}, {and}
  \bibinfo{person}{Daniele Magazzeni}.} \bibinfo{year}{2019}\natexlab{}.
\newblock \showarticletitle{Distilling deep reinforcement learning policies in
  soft decision trees}. In \bibinfo{booktitle}{\emph{Proceedings of the IJCAI
  2019 workshop on explainable artificial intelligence}}.
  \bibinfo{pages}{1--6}.
\newblock


\bibitem[Coronato et~al\mbox{.}(2020)]%
        {coronato2020reinforcement}
\bibfield{author}{\bibinfo{person}{Antonio Coronato}, \bibinfo{person}{Muddasar
  Naeem}, \bibinfo{person}{Giuseppe De~Pietro}, {and} \bibinfo{person}{Giovanni
  Paragliola}.} \bibinfo{year}{2020}\natexlab{}.
\newblock \showarticletitle{Reinforcement learning for intelligent healthcare
  applications: A survey}.
\newblock \bibinfo{journal}{\emph{Artificial Intelligence in Medicine}}
  \bibinfo{volume}{109} (\bibinfo{year}{2020}), \bibinfo{pages}{101964}.
\newblock


\bibitem[Dai et~al\mbox{.}(2022)]%
        {dai2022counterfactual}
\bibfield{author}{\bibinfo{person}{Xinyue Dai}, \bibinfo{person}{Mark~T Keane},
  \bibinfo{person}{Laurence Shalloo}, \bibinfo{person}{Elodie Ruelle}, {and}
  \bibinfo{person}{Ruth~MJ Byrne}.} \bibinfo{year}{2022}\natexlab{}.
\newblock \showarticletitle{Counterfactual explanations for prediction and
  diagnosis in xai}. In \bibinfo{booktitle}{\emph{Proceedings of the 2022
  AAAI/ACM Conference on AI, Ethics, and Society}}. \bibinfo{pages}{215--226}.
\newblock


\bibitem[Dandl et~al\mbox{.}(2020)]%
        {dandl2020multi}
\bibfield{author}{\bibinfo{person}{Susanne Dandl}, \bibinfo{person}{Christoph
  Molnar}, \bibinfo{person}{Martin Binder}, {and} \bibinfo{person}{Bernd
  Bischl}.} \bibinfo{year}{2020}\natexlab{}.
\newblock \showarticletitle{Multi-objective counterfactual explanations}. In
  \bibinfo{booktitle}{\emph{International Conference on Parallel Problem
  Solving from Nature}}. Springer, \bibinfo{pages}{448--469}.
\newblock


\bibitem[Dazeley et~al\mbox{.}(2023)]%
        {broadXAI}
\bibfield{author}{\bibinfo{person}{Richard Dazeley}, \bibinfo{person}{Peter
  Vamplew}, {and} \bibinfo{person}{Francisco Cruz}.}
  \bibinfo{year}{2023}\natexlab{}.
\newblock \showarticletitle{Explainable reinforcement learning for broad-xai: a
  conceptual framework and survey}.
\newblock \bibinfo{journal}{\emph{Neural Computing and Applications}}
  (\bibinfo{year}{2023}), \bibinfo{pages}{1--24}.
\newblock


\bibitem[Dazeley et~al\mbox{.}(2021)]%
        {dazeley2021levels}
\bibfield{author}{\bibinfo{person}{Richard Dazeley}, \bibinfo{person}{Peter
  Vamplew}, \bibinfo{person}{Cameron Foale}, \bibinfo{person}{Charlotte Young},
  \bibinfo{person}{Sunil Aryal}, {and} \bibinfo{person}{Francisco Cruz}.}
  \bibinfo{year}{2021}\natexlab{}.
\newblock \showarticletitle{Levels of explainable artificial intelligence for
  human-aligned conversational explanations}.
\newblock \bibinfo{journal}{\emph{Artificial Intelligence}}
  \bibinfo{volume}{299} (\bibinfo{year}{2021}), \bibinfo{pages}{103525}.
\newblock


\bibitem[Deb et~al\mbox{.}(2002)]%
        {deb2002fast}
\bibfield{author}{\bibinfo{person}{Kalyanmoy Deb}, \bibinfo{person}{Amrit
  Pratap}, \bibinfo{person}{Sameer Agarwal}, {and} \bibinfo{person}{TAMT
  Meyarivan}.} \bibinfo{year}{2002}\natexlab{}.
\newblock \showarticletitle{A fast and elitist multiobjective genetic
  algorithm: NSGA-II}.
\newblock \bibinfo{journal}{\emph{IEEE transactions on evolutionary
  computation}} \bibinfo{volume}{6}, \bibinfo{number}{2}
  (\bibinfo{year}{2002}), \bibinfo{pages}{182--197}.
\newblock


\bibitem[Dethise et~al\mbox{.}(2019)]%
        {dethise2019cracking}
\bibfield{author}{\bibinfo{person}{Arnaud Dethise}, \bibinfo{person}{Marco
  Canini}, {and} \bibinfo{person}{Srikanth Kandula}.}
  \bibinfo{year}{2019}\natexlab{}.
\newblock \showarticletitle{Cracking open the black box: What observations can
  tell us about reinforcement learning agents}. In
  \bibinfo{booktitle}{\emph{Proceedings of the 2019 Workshop on Network Meets
  AI \& ML}}. \bibinfo{pages}{29--36}.
\newblock


\bibitem[Dulac-Arnold et~al\mbox{.}(2019)]%
        {dulac2019challenges}
\bibfield{author}{\bibinfo{person}{Gabriel Dulac-Arnold},
  \bibinfo{person}{Daniel Mankowitz}, {and} \bibinfo{person}{Todd Hester}.}
  \bibinfo{year}{2019}\natexlab{}.
\newblock \showarticletitle{Challenges of real-world reinforcement learning}.
\newblock \bibinfo{journal}{\emph{arXiv preprint arXiv:1904.12901}}
  (\bibinfo{year}{2019}).
\newblock


\bibitem[Ferrante et~al\mbox{.}(2013)]%
        {ferrante2013improving}
\bibfield{author}{\bibinfo{person}{Donatella Ferrante},
  \bibinfo{person}{Vittorio Girotto}, \bibinfo{person}{Marta Straga}, {and}
  \bibinfo{person}{Clare Walsh}.} \bibinfo{year}{2013}\natexlab{}.
\newblock \showarticletitle{Improving the past and the future: a temporal
  asymmetry in hypothetical thinking.}
\newblock \bibinfo{journal}{\emph{Journal of Experimental Psychology: General}}
  \bibinfo{volume}{142}, \bibinfo{number}{1} (\bibinfo{year}{2013}),
  \bibinfo{pages}{23}.
\newblock


\bibitem[Frosst and Hinton(2017)]%
        {frosst2017distilling}
\bibfield{author}{\bibinfo{person}{Nicholas Frosst} {and}
  \bibinfo{person}{Geoffrey Hinton}.} \bibinfo{year}{2017}\natexlab{}.
\newblock \showarticletitle{Distilling a neural network into a soft decision
  tree}.
\newblock \bibinfo{journal}{\emph{arXiv preprint arXiv:1711.09784}}
  (\bibinfo{year}{2017}).
\newblock


\bibitem[Gajcin and Dusparic(2022)]%
        {gajcin2022reccover}
\bibfield{author}{\bibinfo{person}{Jasmina Gajcin} {and} \bibinfo{person}{Ivana
  Dusparic}.} \bibinfo{year}{2022}\natexlab{}.
\newblock \showarticletitle{ReCCoVER: Detecting Causal Confusion for
  Explainable Reinforcement Learning}.
\newblock \bibinfo{journal}{\emph{arXiv preprint arXiv:2203.11211}}
  (\bibinfo{year}{2022}).
\newblock


\bibitem[Gajcin et~al\mbox{.}(2021)]%
        {gajcin2021contrastive}
\bibfield{author}{\bibinfo{person}{Jasmina Gajcin}, \bibinfo{person}{Rahul
  Nair}, \bibinfo{person}{Tejaswini Pedapati}, \bibinfo{person}{Radu
  Marinescu}, \bibinfo{person}{Elizabeth Daly}, {and} \bibinfo{person}{Ivana
  Dusparic}.} \bibinfo{year}{2021}\natexlab{}.
\newblock \showarticletitle{Contrastive Explanations for Comparing Preferences
  of Reinforcement Learning Agents}.
\newblock \bibinfo{journal}{\emph{arXiv preprint arXiv:2112.09462}}
  (\bibinfo{year}{2021}).
\newblock


\bibitem[Garc{\i}a and Fern{\'a}ndez(2015)]%
        {garcia2015comprehensive}
\bibfield{author}{\bibinfo{person}{Javier Garc{\i}a} {and}
  \bibinfo{person}{Fernando Fern{\'a}ndez}.} \bibinfo{year}{2015}\natexlab{}.
\newblock \showarticletitle{A comprehensive survey on safe reinforcement
  learning}.
\newblock \bibinfo{journal}{\emph{Journal of Machine Learning Research}}
  \bibinfo{volume}{16}, \bibinfo{number}{1} (\bibinfo{year}{2015}),
  \bibinfo{pages}{1437--1480}.
\newblock


\bibitem[Gleave et~al\mbox{.}(2020)]%
        {gleave2020quantifying}
\bibfield{author}{\bibinfo{person}{Adam Gleave}, \bibinfo{person}{Michael
  Dennis}, \bibinfo{person}{Shane Legg}, \bibinfo{person}{Stuart Russell},
  {and} \bibinfo{person}{Jan Leike}.} \bibinfo{year}{2020}\natexlab{}.
\newblock \showarticletitle{Quantifying differences in reward functions}.
\newblock \bibinfo{journal}{\emph{arXiv preprint arXiv:2006.13900}}
  (\bibinfo{year}{2020}).
\newblock


\bibitem[Goodman and Flaxman(2017)]%
        {goodman2017european}
\bibfield{author}{\bibinfo{person}{Bryce Goodman} {and} \bibinfo{person}{Seth
  Flaxman}.} \bibinfo{year}{2017}\natexlab{}.
\newblock \showarticletitle{European Union regulations on algorithmic
  decision-making and a “right to explanation”}.
\newblock \bibinfo{journal}{\emph{AI magazine}} \bibinfo{volume}{38},
  \bibinfo{number}{3} (\bibinfo{year}{2017}), \bibinfo{pages}{50--57}.
\newblock


\bibitem[Greydanus et~al\mbox{.}(2018)]%
        {greydanus2018visualizing}
\bibfield{author}{\bibinfo{person}{Samuel Greydanus}, \bibinfo{person}{Anurag
  Koul}, \bibinfo{person}{Jonathan Dodge}, {and} \bibinfo{person}{Alan Fern}.}
  \bibinfo{year}{2018}\natexlab{}.
\newblock \showarticletitle{Visualizing and understanding atari agents}. In
  \bibinfo{booktitle}{\emph{International Conference on Machine Learning}}.
  PMLR, \bibinfo{pages}{1792--1801}.
\newblock


\bibitem[Guidotti(2022)]%
        {guidotti2022counterfactual}
\bibfield{author}{\bibinfo{person}{Riccardo Guidotti}.}
  \bibinfo{year}{2022}\natexlab{}.
\newblock \showarticletitle{Counterfactual explanations and how to find them:
  literature review and benchmarking}.
\newblock \bibinfo{journal}{\emph{Data Mining and Knowledge Discovery}}
  (\bibinfo{year}{2022}), \bibinfo{pages}{1--55}.
\newblock


\bibitem[Guidotti et~al\mbox{.}(2019)]%
        {1283}
\bibfield{author}{\bibinfo{person}{Riccardo Guidotti}, \bibinfo{person}{Anna
  Monreale}, \bibinfo{person}{Fosca Giannotti}, \bibinfo{person}{Dino
  Pedreschi}, \bibinfo{person}{Salvatore Ruggieri}, {and}
  \bibinfo{person}{Franco Turini}.} \bibinfo{year}{2019}\natexlab{}.
\newblock \showarticletitle{Factual and Counterfactual Explanations for Black
  Box Decision Making}.
\newblock \bibinfo{journal}{\emph{IEEE Intelligent Systems}}
  (\bibinfo{year}{2019}).
\newblock
\urldef\tempurl%
\url{https://doi.org/10.1109/MIS.2019.2957223}
\showDOI{\tempurl}


\bibitem[Harutyunyan et~al\mbox{.}(2019)]%
        {harutyunyan2019hindsight}
\bibfield{author}{\bibinfo{person}{Anna Harutyunyan}, \bibinfo{person}{Will
  Dabney}, \bibinfo{person}{Thomas Mesnard}, \bibinfo{person}{Mohammad
  Gheshlaghi~Azar}, \bibinfo{person}{Bilal Piot}, \bibinfo{person}{Nicolas
  Heess}, \bibinfo{person}{Hado~P van Hasselt}, \bibinfo{person}{Gregory
  Wayne}, \bibinfo{person}{Satinder Singh}, \bibinfo{person}{Doina Precup},
  {et~al\mbox{.}}} \bibinfo{year}{2019}\natexlab{}.
\newblock \showarticletitle{Hindsight credit assignment}.
\newblock \bibinfo{journal}{\emph{Advances in neural information processing
  systems}}  \bibinfo{volume}{32} (\bibinfo{year}{2019}).
\newblock


\bibitem[Heuillet et~al\mbox{.}(2021)]%
        {heuillet2021explainability}
\bibfield{author}{\bibinfo{person}{Alexandre Heuillet}, \bibinfo{person}{Fabien
  Couthouis}, {and} \bibinfo{person}{Natalia D{\'\i}az-Rodr{\'\i}guez}.}
  \bibinfo{year}{2021}\natexlab{}.
\newblock \showarticletitle{Explainability in deep reinforcement learning}.
\newblock \bibinfo{journal}{\emph{Knowledge-Based Systems}}
  \bibinfo{volume}{214} (\bibinfo{year}{2021}), \bibinfo{pages}{106685}.
\newblock


\bibitem[Huber et~al\mbox{.}(2023)]%
        {huber2023ganterfactual}
\bibfield{author}{\bibinfo{person}{Tobias Huber}, \bibinfo{person}{Maximilian
  Demmler}, \bibinfo{person}{Silvan Mertes}, \bibinfo{person}{Matthew~L Olson},
  {and} \bibinfo{person}{Elisabeth Andr{\'e}}.}
  \bibinfo{year}{2023}\natexlab{}.
\newblock \showarticletitle{GANterfactual-RL: Understanding Reinforcement
  Learning Agents' Strategies through Visual Counterfactual Explanations}.
\newblock \bibinfo{journal}{\emph{arXiv preprint arXiv:2302.12689}}
  (\bibinfo{year}{2023}).
\newblock


\bibitem[Huber et~al\mbox{.}(2021)]%
        {huber2021benchmarking}
\bibfield{author}{\bibinfo{person}{Tobias Huber}, \bibinfo{person}{Benedikt
  Limmer}, {and} \bibinfo{person}{Elisabeth Andr{\'e}}.}
  \bibinfo{year}{2021}\natexlab{}.
\newblock \showarticletitle{Benchmarking perturbation-based saliency maps for
  explaining deep reinforcement learning agents}.
\newblock \bibinfo{journal}{\emph{arXiv e-prints}} (\bibinfo{year}{2021}),
  \bibinfo{pages}{arXiv--2101}.
\newblock


\bibitem[Jentzsch et~al\mbox{.}(2019)]%
        {jentzsch2019semantics}
\bibfield{author}{\bibinfo{person}{Sophie Jentzsch}, \bibinfo{person}{Patrick
  Schramowski}, \bibinfo{person}{Constantin Rothkopf}, {and}
  \bibinfo{person}{Kristian Kersting}.} \bibinfo{year}{2019}\natexlab{}.
\newblock \showarticletitle{Semantics derived automatically from language
  corpora contain human-like moral choices}. In
  \bibinfo{booktitle}{\emph{Proceedings of the 2019 AAAI/ACM Conference on AI,
  Ethics, and Society}}. \bibinfo{pages}{37--44}.
\newblock


\bibitem[Joshi et~al\mbox{.}(2019)]%
        {joshi2019towards}
\bibfield{author}{\bibinfo{person}{Shalmali Joshi}, \bibinfo{person}{Oluwasanmi
  Koyejo}, \bibinfo{person}{Warut Vijitbenjaronk}, \bibinfo{person}{Been Kim},
  {and} \bibinfo{person}{Joydeep Ghosh}.} \bibinfo{year}{2019}\natexlab{}.
\newblock \showarticletitle{Towards realistic individual recourse and
  actionable explanations in black-box decision making systems}.
\newblock \bibinfo{journal}{\emph{arXiv preprint arXiv:1907.09615}}
  (\bibinfo{year}{2019}).
\newblock


\bibitem[Juozapaitis et~al\mbox{.}(2019)]%
        {juozapaitis2019explainable}
\bibfield{author}{\bibinfo{person}{Zoe Juozapaitis}, \bibinfo{person}{Anurag
  Koul}, \bibinfo{person}{Alan Fern}, \bibinfo{person}{Martin Erwig}, {and}
  \bibinfo{person}{Finale Doshi-Velez}.} \bibinfo{year}{2019}\natexlab{}.
\newblock \showarticletitle{Explainable reinforcement learning via reward
  decomposition}. In \bibinfo{booktitle}{\emph{IJCAI/ECAI Workshop on
  Explainable Artificial Intelligence}}.
\newblock


\bibitem[Karakovskiy and Togelius(2012)]%
        {karakovskiy2012mario}
\bibfield{author}{\bibinfo{person}{Sergey Karakovskiy} {and}
  \bibinfo{person}{Julian Togelius}.} \bibinfo{year}{2012}\natexlab{}.
\newblock \showarticletitle{The mario ai benchmark and competitions}.
\newblock \bibinfo{journal}{\emph{IEEE Transactions on Computational
  Intelligence and AI in Games}} \bibinfo{volume}{4}, \bibinfo{number}{1}
  (\bibinfo{year}{2012}), \bibinfo{pages}{55--67}.
\newblock


\bibitem[Karimi et~al\mbox{.}(2020a)]%
        {karimi2020survey}
\bibfield{author}{\bibinfo{person}{Amir-Hossein Karimi},
  \bibinfo{person}{Gilles Barthe}, \bibinfo{person}{Bernhard Sch{\"o}lkopf},
  {and} \bibinfo{person}{Isabel Valera}.} \bibinfo{year}{2020}\natexlab{a}.
\newblock \showarticletitle{A survey of algorithmic recourse: definitions,
  formulations, solutions, and prospects}.
\newblock \bibinfo{journal}{\emph{arXiv preprint arXiv:2010.04050}}
  (\bibinfo{year}{2020}).
\newblock


\bibitem[Karimi et~al\mbox{.}(2021)]%
        {karimi2021algorithmic}
\bibfield{author}{\bibinfo{person}{Amir-Hossein Karimi},
  \bibinfo{person}{Bernhard Sch{\"o}lkopf}, {and} \bibinfo{person}{Isabel
  Valera}.} \bibinfo{year}{2021}\natexlab{}.
\newblock \showarticletitle{Algorithmic recourse: from counterfactual
  explanations to interventions}. In \bibinfo{booktitle}{\emph{Proceedings of
  the 2021 ACM Conference on Fairness, Accountability, and Transparency}}.
  \bibinfo{pages}{353--362}.
\newblock


\bibitem[Karimi et~al\mbox{.}(2020b)]%
        {karimi2020algorithmic}
\bibfield{author}{\bibinfo{person}{Amir-Hossein Karimi},
  \bibinfo{person}{Julius Von~K{\"u}gelgen}, \bibinfo{person}{Bernhard
  Sch{\"o}lkopf}, {and} \bibinfo{person}{Isabel Valera}.}
  \bibinfo{year}{2020}\natexlab{b}.
\newblock \showarticletitle{Algorithmic recourse under imperfect causal
  knowledge: a probabilistic approach}.
\newblock \bibinfo{journal}{\emph{Advances in Neural Information Processing
  Systems}}  \bibinfo{volume}{33} (\bibinfo{year}{2020}),
  \bibinfo{pages}{265--277}.
\newblock


\bibitem[Kiran et~al\mbox{.}(2021)]%
        {kiran2021deep}
\bibfield{author}{\bibinfo{person}{B~Ravi Kiran}, \bibinfo{person}{Ibrahim
  Sobh}, \bibinfo{person}{Victor Talpaert}, \bibinfo{person}{Patrick Mannion},
  \bibinfo{person}{Ahmad~A Al~Sallab}, \bibinfo{person}{Senthil Yogamani},
  {and} \bibinfo{person}{Patrick P{\'e}rez}.} \bibinfo{year}{2021}\natexlab{}.
\newblock \showarticletitle{Deep reinforcement learning for autonomous driving:
  A survey}.
\newblock \bibinfo{journal}{\emph{IEEE Transactions on Intelligent
  Transportation Systems}} (\bibinfo{year}{2021}).
\newblock


\bibitem[Knight(2018)]%
        {knight_2018}
\bibfield{author}{\bibinfo{person}{Will Knight}.}
  \bibinfo{year}{2018}\natexlab{}.
\newblock \bibinfo{title}{What Uber’s fatal accident could mean for the
  autonomous-car industry}.
\newblock
\newblock
\urldef\tempurl%
\url{technologyreview.com/2018/03/19/241022/what-ubers-fatal-accident-could-mean-for-the-autonomous-car-industry}
\showURL{%
\tempurl}


\bibitem[Kober et~al\mbox{.}(2013)]%
        {kober2013reinforcement}
\bibfield{author}{\bibinfo{person}{Jens Kober}, \bibinfo{person}{J~Andrew
  Bagnell}, {and} \bibinfo{person}{Jan Peters}.}
  \bibinfo{year}{2013}\natexlab{}.
\newblock \showarticletitle{Reinforcement learning in robotics: A survey}.
\newblock \bibinfo{journal}{\emph{The International Journal of Robotics
  Research}} \bibinfo{volume}{32}, \bibinfo{number}{11} (\bibinfo{year}{2013}),
  \bibinfo{pages}{1238--1274}.
\newblock


\bibitem[Kundu(2021)]%
        {kundu2021ai}
\bibfield{author}{\bibinfo{person}{Shinjini Kundu}.}
  \bibinfo{year}{2021}\natexlab{}.
\newblock \showarticletitle{AI in medicine must be explainable}.
\newblock \bibinfo{journal}{\emph{Nature medicine}} \bibinfo{volume}{27},
  \bibinfo{number}{8} (\bibinfo{year}{2021}), \bibinfo{pages}{1328--1328}.
\newblock


\bibitem[Lakkaraju et~al\mbox{.}(2017)]%
        {lakkaraju2017interpretable}
\bibfield{author}{\bibinfo{person}{Himabindu Lakkaraju}, \bibinfo{person}{Ece
  Kamar}, \bibinfo{person}{Rich Caruana}, {and} \bibinfo{person}{Jure
  Leskovec}.} \bibinfo{year}{2017}\natexlab{}.
\newblock \showarticletitle{Interpretable \& explorable approximations of black
  box models}.
\newblock \bibinfo{journal}{\emph{arXiv preprint arXiv:1707.01154}}
  (\bibinfo{year}{2017}).
\newblock


\bibitem[Landajuela et~al\mbox{.}(2021)]%
        {landajuela2021discovering}
\bibfield{author}{\bibinfo{person}{Mikel Landajuela},
  \bibinfo{person}{Brenden~K Petersen}, \bibinfo{person}{Sookyung Kim},
  \bibinfo{person}{Claudio~P Santiago}, \bibinfo{person}{Ruben Glatt},
  \bibinfo{person}{Nathan Mundhenk}, \bibinfo{person}{Jacob~F Pettit}, {and}
  \bibinfo{person}{Daniel Faissol}.} \bibinfo{year}{2021}\natexlab{}.
\newblock \showarticletitle{Discovering symbolic policies with deep
  reinforcement learning}. In \bibinfo{booktitle}{\emph{International
  Conference on Machine Learning}}. PMLR, \bibinfo{pages}{5979--5989}.
\newblock


\bibitem[Laugel et~al\mbox{.}(2017)]%
        {laugel2017inverse}
\bibfield{author}{\bibinfo{person}{Thibault Laugel},
  \bibinfo{person}{Marie-Jeanne Lesot}, \bibinfo{person}{Christophe Marsala},
  \bibinfo{person}{Xavier Renard}, {and} \bibinfo{person}{Marcin Detyniecki}.}
  \bibinfo{year}{2017}\natexlab{}.
\newblock \showarticletitle{Inverse classification for comparison-based
  interpretability in machine learning}.
\newblock \bibinfo{journal}{\emph{arXiv preprint arXiv:1712.08443}}
  (\bibinfo{year}{2017}).
\newblock


\bibitem[Li(2019)]%
        {li2019reinforcement}
\bibfield{author}{\bibinfo{person}{Yuxi Li}.} \bibinfo{year}{2019}\natexlab{}.
\newblock \showarticletitle{Reinforcement learning applications}.
\newblock \bibinfo{journal}{\emph{arXiv preprint arXiv:1908.06973}}
  (\bibinfo{year}{2019}).
\newblock


\bibitem[Lillicrap et~al\mbox{.}(2015)]%
        {lillicrap2015continuous}
\bibfield{author}{\bibinfo{person}{Timothy~P Lillicrap},
  \bibinfo{person}{Jonathan~J Hunt}, \bibinfo{person}{Alexander Pritzel},
  \bibinfo{person}{Nicolas Heess}, \bibinfo{person}{Tom Erez},
  \bibinfo{person}{Yuval Tassa}, \bibinfo{person}{David Silver}, {and}
  \bibinfo{person}{Daan Wierstra}.} \bibinfo{year}{2015}\natexlab{}.
\newblock \showarticletitle{Continuous control with deep reinforcement
  learning}.
\newblock \bibinfo{journal}{\emph{arXiv preprint arXiv:1509.02971}}
  (\bibinfo{year}{2015}).
\newblock


\bibitem[Lipton(1990)]%
        {lipton1990contrastive}
\bibfield{author}{\bibinfo{person}{Peter Lipton}.}
  \bibinfo{year}{1990}\natexlab{}.
\newblock \showarticletitle{Contrastive explanation}.
\newblock \bibinfo{journal}{\emph{Royal Institute of Philosophy Supplements}}
  \bibinfo{volume}{27} (\bibinfo{year}{1990}), \bibinfo{pages}{247--266}.
\newblock


\bibitem[Liu et~al\mbox{.}(2014)]%
        {liu2014multiobjective}
\bibfield{author}{\bibinfo{person}{Chunming Liu}, \bibinfo{person}{Xin Xu},
  {and} \bibinfo{person}{Dewen Hu}.} \bibinfo{year}{2014}\natexlab{}.
\newblock \showarticletitle{Multiobjective reinforcement learning: A
  comprehensive overview}.
\newblock \bibinfo{journal}{\emph{IEEE Transactions on Systems, Man, and
  Cybernetics: Systems}} \bibinfo{volume}{45}, \bibinfo{number}{3}
  (\bibinfo{year}{2014}), \bibinfo{pages}{385--398}.
\newblock


\bibitem[Liu et~al\mbox{.}(2018)]%
        {liu2018toward}
\bibfield{author}{\bibinfo{person}{Guiliang Liu}, \bibinfo{person}{Oliver
  Schulte}, \bibinfo{person}{Wang Zhu}, {and} \bibinfo{person}{Qingcan Li}.}
  \bibinfo{year}{2018}\natexlab{}.
\newblock \showarticletitle{Toward interpretable deep reinforcement learning
  with linear model u-trees}. In \bibinfo{booktitle}{\emph{Joint European
  Conference on Machine Learning and Knowledge Discovery in Databases}}.
  Springer, \bibinfo{pages}{414--429}.
\newblock


\bibitem[Looveren and Klaise(2021)]%
        {looveren2021interpretable}
\bibfield{author}{\bibinfo{person}{Arnaud~Van Looveren} {and}
  \bibinfo{person}{Janis Klaise}.} \bibinfo{year}{2021}\natexlab{}.
\newblock \showarticletitle{Interpretable counterfactual explanations guided by
  prototypes}. In \bibinfo{booktitle}{\emph{Joint European Conference on
  Machine Learning and Knowledge Discovery in Databases}}. Springer,
  \bibinfo{pages}{650--665}.
\newblock


\bibitem[Luckett et~al\mbox{.}(2019)]%
        {luckett2019estimating}
\bibfield{author}{\bibinfo{person}{Daniel~J Luckett}, \bibinfo{person}{Eric~B
  Laber}, \bibinfo{person}{Anna~R Kahkoska}, \bibinfo{person}{David~M Maahs},
  \bibinfo{person}{Elizabeth Mayer-Davis}, {and} \bibinfo{person}{Michael~R
  Kosorok}.} \bibinfo{year}{2019}\natexlab{}.
\newblock \showarticletitle{Estimating dynamic treatment regimes in mobile
  health using v-learning}.
\newblock \bibinfo{journal}{\emph{J. Amer. Statist. Assoc.}}
  (\bibinfo{year}{2019}).
\newblock


\bibitem[Lundberg and Lee(2017)]%
        {lundberg2017unified}
\bibfield{author}{\bibinfo{person}{Scott~M Lundberg} {and}
  \bibinfo{person}{Su-In Lee}.} \bibinfo{year}{2017}\natexlab{}.
\newblock \showarticletitle{A unified approach to interpreting model
  predictions}. In \bibinfo{booktitle}{\emph{Proceedings of the 31st
  international conference on neural information processing systems}}.
  \bibinfo{pages}{4768--4777}.
\newblock


\bibitem[Madumal et~al\mbox{.}(2020)]%
        {madumal2020explainable}
\bibfield{author}{\bibinfo{person}{Prashan Madumal}, \bibinfo{person}{Tim
  Miller}, \bibinfo{person}{Liz Sonenberg}, {and} \bibinfo{person}{Frank
  Vetere}.} \bibinfo{year}{2020}\natexlab{}.
\newblock \showarticletitle{Explainable reinforcement learning through a causal
  lens}. In \bibinfo{booktitle}{\emph{Proceedings of the AAAI Conference on
  Artificial Intelligence}}, Vol.~\bibinfo{volume}{34}.
  \bibinfo{pages}{2493--2500}.
\newblock


\bibitem[Mahajan et~al\mbox{.}(2019)]%
        {mahajan2019preserving}
\bibfield{author}{\bibinfo{person}{Divyat Mahajan}, \bibinfo{person}{Chenhao
  Tan}, {and} \bibinfo{person}{Amit Sharma}.} \bibinfo{year}{2019}\natexlab{}.
\newblock \showarticletitle{Preserving causal constraints in counterfactual
  explanations for machine learning classifiers}.
\newblock \bibinfo{journal}{\emph{arXiv preprint arXiv:1912.03277}}
  (\bibinfo{year}{2019}).
\newblock


\bibitem[Miller(2019)]%
        {miller2019explanation}
\bibfield{author}{\bibinfo{person}{Tim Miller}.}
  \bibinfo{year}{2019}\natexlab{}.
\newblock \showarticletitle{Explanation in artificial intelligence: Insights
  from the social sciences}.
\newblock \bibinfo{journal}{\emph{Artificial intelligence}}
  \bibinfo{volume}{267} (\bibinfo{year}{2019}), \bibinfo{pages}{1--38}.
\newblock


\bibitem[Mnih et~al\mbox{.}(2013)]%
        {mnih2013playing}
\bibfield{author}{\bibinfo{person}{Volodymyr Mnih}, \bibinfo{person}{Koray
  Kavukcuoglu}, \bibinfo{person}{David Silver}, \bibinfo{person}{Alex Graves},
  \bibinfo{person}{Ioannis Antonoglou}, \bibinfo{person}{Daan Wierstra}, {and}
  \bibinfo{person}{Martin Riedmiller}.} \bibinfo{year}{2013}\natexlab{}.
\newblock \showarticletitle{Playing atari with deep reinforcement learning}.
\newblock \bibinfo{journal}{\emph{arXiv preprint arXiv:1312.5602}}
  (\bibinfo{year}{2013}).
\newblock


\bibitem[Molnar(2019)]%
        {molnar2019}
\bibfield{author}{\bibinfo{person}{Christoph Molnar}.}
  \bibinfo{year}{2019}\natexlab{}.
\newblock \bibinfo{booktitle}{\emph{Interpretable Machine Learning}}.
\newblock


\bibitem[Mothilal et~al\mbox{.}(2020)]%
        {mothilal2020explaining}
\bibfield{author}{\bibinfo{person}{Ramaravind~K Mothilal},
  \bibinfo{person}{Amit Sharma}, {and} \bibinfo{person}{Chenhao Tan}.}
  \bibinfo{year}{2020}\natexlab{}.
\newblock \showarticletitle{Explaining machine learning classifiers through
  diverse counterfactual explanations}. In
  \bibinfo{booktitle}{\emph{Proceedings of the 2020 Conference on Fairness,
  Accountability, and Transparency}}. \bibinfo{pages}{607--617}.
\newblock


\bibitem[Olson et~al\mbox{.}(2019)]%
        {olson2019counterfactual}
\bibfield{author}{\bibinfo{person}{Matthew~L Olson}, \bibinfo{person}{Lawrence
  Neal}, \bibinfo{person}{Fuxin Li}, {and} \bibinfo{person}{Weng-Keen Wong}.}
  \bibinfo{year}{2019}\natexlab{}.
\newblock \showarticletitle{Counterfactual states for atari agents via
  generative deep learning}.
\newblock \bibinfo{journal}{\emph{arXiv preprint arXiv:1909.12969}}
  (\bibinfo{year}{2019}).
\newblock


\bibitem[Pan et~al\mbox{.}(2022)]%
        {pan2022effects}
\bibfield{author}{\bibinfo{person}{Alexander Pan}, \bibinfo{person}{Kush
  Bhatia}, {and} \bibinfo{person}{Jacob Steinhardt}.}
  \bibinfo{year}{2022}\natexlab{}.
\newblock \showarticletitle{The effects of reward misspecification: Mapping and
  mitigating misaligned models}.
\newblock \bibinfo{journal}{\emph{arXiv preprint arXiv:2201.03544}}
  (\bibinfo{year}{2022}).
\newblock


\bibitem[Pearl and Mackenzie(2018)]%
        {pearl2018book}
\bibfield{author}{\bibinfo{person}{Judea Pearl} {and} \bibinfo{person}{Dana
  Mackenzie}.} \bibinfo{year}{2018}\natexlab{}.
\newblock \bibinfo{booktitle}{\emph{The book of why: the new science of cause
  and effect}}.
\newblock \bibinfo{publisher}{Basic books}.
\newblock


\bibitem[Peng et~al\mbox{.}(2018)]%
        {peng2018deepmimic}
\bibfield{author}{\bibinfo{person}{Xue~Bin Peng}, \bibinfo{person}{Pieter
  Abbeel}, \bibinfo{person}{Sergey Levine}, {and} \bibinfo{person}{Michiel
  van~de Panne}.} \bibinfo{year}{2018}\natexlab{}.
\newblock \showarticletitle{Deepmimic: Example-guided deep reinforcement
  learning of physics-based character skills}.
\newblock \bibinfo{journal}{\emph{ACM Transactions on Graphics (TOG)}}
  \bibinfo{volume}{37}, \bibinfo{number}{4} (\bibinfo{year}{2018}),
  \bibinfo{pages}{1--14}.
\newblock


\bibitem[Pouyanfar et~al\mbox{.}(2018)]%
        {pouyanfar2018survey}
\bibfield{author}{\bibinfo{person}{Samira Pouyanfar}, \bibinfo{person}{Saad
  Sadiq}, \bibinfo{person}{Yilin Yan}, \bibinfo{person}{Haiman Tian},
  \bibinfo{person}{Yudong Tao}, \bibinfo{person}{Maria~Presa Reyes},
  \bibinfo{person}{Mei-Ling Shyu}, \bibinfo{person}{Shu-Ching Chen}, {and}
  \bibinfo{person}{Sundaraja~S Iyengar}.} \bibinfo{year}{2018}\natexlab{}.
\newblock \showarticletitle{A survey on deep learning: Algorithms, techniques,
  and applications}.
\newblock \bibinfo{journal}{\emph{ACM Computing Surveys (CSUR)}}
  \bibinfo{volume}{51}, \bibinfo{number}{5} (\bibinfo{year}{2018}),
  \bibinfo{pages}{1--36}.
\newblock


\bibitem[Poyiadzi et~al\mbox{.}(2020)]%
        {poyiadzi2020face}
\bibfield{author}{\bibinfo{person}{Rafael Poyiadzi}, \bibinfo{person}{Kacper
  Sokol}, \bibinfo{person}{Raul Santos-Rodriguez}, \bibinfo{person}{Tijl
  De~Bie}, {and} \bibinfo{person}{Peter Flach}.}
  \bibinfo{year}{2020}\natexlab{}.
\newblock \showarticletitle{FACE: feasible and actionable counterfactual
  explanations}. In \bibinfo{booktitle}{\emph{Proceedings of the AAAI/ACM
  Conference on AI, Ethics, and Society}}. \bibinfo{pages}{344--350}.
\newblock


\bibitem[Puiutta and Veith(2020)]%
        {puiutta2020explainable}
\bibfield{author}{\bibinfo{person}{Erika Puiutta} {and} \bibinfo{person}{Eric
  Veith}.} \bibinfo{year}{2020}\natexlab{}.
\newblock \showarticletitle{Explainable reinforcement learning: A survey}. In
  \bibinfo{booktitle}{\emph{International cross-domain conference for machine
  learning and knowledge extraction}}. Springer, \bibinfo{pages}{77--95}.
\newblock


\bibitem[Puri et~al\mbox{.}(2019)]%
        {puri2019explain}
\bibfield{author}{\bibinfo{person}{Nikaash Puri}, \bibinfo{person}{Sukriti
  Verma}, \bibinfo{person}{Piyush Gupta}, \bibinfo{person}{Dhruv Kayastha},
  \bibinfo{person}{Shripad Deshmukh}, \bibinfo{person}{Balaji Krishnamurthy},
  {and} \bibinfo{person}{Sameer Singh}.} \bibinfo{year}{2019}\natexlab{}.
\newblock \showarticletitle{Explain your move: Understanding agent actions
  using specific and relevant feature attribution}.
\newblock \bibinfo{journal}{\emph{arXiv preprint arXiv:1912.12191}}
  (\bibinfo{year}{2019}).
\newblock


\bibitem[Riachi et~al\mbox{.}(2021)]%
        {riachi2021challenges}
\bibfield{author}{\bibinfo{person}{Elsa Riachi}, \bibinfo{person}{Muhammad
  Mamdani}, \bibinfo{person}{Michael Fralick}, {and} \bibinfo{person}{Frank
  Rudzicz}.} \bibinfo{year}{2021}\natexlab{}.
\newblock \showarticletitle{Challenges for reinforcement learning in
  healthcare}.
\newblock \bibinfo{journal}{\emph{arXiv preprint arXiv:2103.05612}}
  (\bibinfo{year}{2021}).
\newblock


\bibitem[Ribeiro et~al\mbox{.}(2016)]%
        {ribeiro2016should}
\bibfield{author}{\bibinfo{person}{Marco~Tulio Ribeiro},
  \bibinfo{person}{Sameer Singh}, {and} \bibinfo{person}{Carlos Guestrin}.}
  \bibinfo{year}{2016}\natexlab{}.
\newblock \showarticletitle{" Why should i trust you?" Explaining the
  predictions of any classifier}. In \bibinfo{booktitle}{\emph{Proceedings of
  the 22nd ACM SIGKDD international conference on knowledge discovery and data
  mining}}. \bibinfo{pages}{1135--1144}.
\newblock


\bibitem[Samoilescu et~al\mbox{.}(2021)]%
        {samoilescu2021model}
\bibfield{author}{\bibinfo{person}{Robert-Florian Samoilescu},
  \bibinfo{person}{Arnaud Van~Looveren}, {and} \bibinfo{person}{Janis Klaise}.}
  \bibinfo{year}{2021}\natexlab{}.
\newblock \showarticletitle{Model-agnostic and Scalable Counterfactual
  Explanations via Reinforcement Learning}.
\newblock \bibinfo{journal}{\emph{arXiv preprint arXiv:2106.02597}}
  (\bibinfo{year}{2021}).
\newblock


\bibitem[Sequeira and Gervasio(2020)]%
        {sequeira2020interestingness}
\bibfield{author}{\bibinfo{person}{Pedro Sequeira} {and}
  \bibinfo{person}{Melinda Gervasio}.} \bibinfo{year}{2020}\natexlab{}.
\newblock \showarticletitle{Interestingness elements for explainable
  reinforcement learning: Understanding agents' capabilities and limitations}.
\newblock \bibinfo{journal}{\emph{Artificial Intelligence}}
  \bibinfo{volume}{288} (\bibinfo{year}{2020}), \bibinfo{pages}{103367}.
\newblock


\bibitem[Sharma et~al\mbox{.}(2019)]%
        {sharma2019certifai}
\bibfield{author}{\bibinfo{person}{Shubham Sharma}, \bibinfo{person}{Jette
  Henderson}, {and} \bibinfo{person}{Joydeep Ghosh}.}
  \bibinfo{year}{2019}\natexlab{}.
\newblock \showarticletitle{Certifai: Counterfactual explanations for
  robustness, transparency, interpretability, and fairness of artificial
  intelligence models}.
\newblock \bibinfo{journal}{\emph{arXiv preprint arXiv:1905.07857}}
  (\bibinfo{year}{2019}).
\newblock


\bibitem[Simonyan et~al\mbox{.}(2013)]%
        {simonyan2013deep}
\bibfield{author}{\bibinfo{person}{Karen Simonyan}, \bibinfo{person}{Andrea
  Vedaldi}, {and} \bibinfo{person}{Andrew Zisserman}.}
  \bibinfo{year}{2013}\natexlab{}.
\newblock \showarticletitle{Deep inside convolutional networks: Visualising
  image classification models and saliency maps}.
\newblock \bibinfo{journal}{\emph{arXiv preprint arXiv:1312.6034}}
  (\bibinfo{year}{2013}).
\newblock


\bibitem[Sokol and Flach(2019)]%
        {sokol2019counterfactual}
\bibfield{author}{\bibinfo{person}{Kacper Sokol} {and} \bibinfo{person}{Peter~A
  Flach}.} \bibinfo{year}{2019}\natexlab{}.
\newblock \showarticletitle{Counterfactual explanations of machine learning
  predictions: opportunities and challenges for AI safety}.
\newblock \bibinfo{journal}{\emph{SafeAI@ AAAI}} (\bibinfo{year}{2019}).
\newblock


\bibitem[Stepin et~al\mbox{.}(2021)]%
        {stepin2021survey}
\bibfield{author}{\bibinfo{person}{Ilia Stepin}, \bibinfo{person}{Jose~M
  Alonso}, \bibinfo{person}{Alejandro Catala}, {and}
  \bibinfo{person}{Mart{\'\i}n Pereira-Fari{\~n}a}.}
  \bibinfo{year}{2021}\natexlab{}.
\newblock \showarticletitle{A survey of contrastive and counterfactual
  explanation generation methods for explainable artificial intelligence}.
\newblock \bibinfo{journal}{\emph{IEEE Access}}  \bibinfo{volume}{9}
  (\bibinfo{year}{2021}), \bibinfo{pages}{11974--12001}.
\newblock


\bibitem[{\v{S}}trumbelj and Kononenko(2014)]%
        {vstrumbelj2014explaining}
\bibfield{author}{\bibinfo{person}{Erik {\v{S}}trumbelj} {and}
  \bibinfo{person}{Igor Kononenko}.} \bibinfo{year}{2014}\natexlab{}.
\newblock \showarticletitle{Explaining prediction models and individual
  predictions with feature contributions}.
\newblock \bibinfo{journal}{\emph{Knowledge and information systems}}
  \bibinfo{volume}{41}, \bibinfo{number}{3} (\bibinfo{year}{2014}),
  \bibinfo{pages}{647--665}.
\newblock


\bibitem[Sukkerd et~al\mbox{.}(2020)]%
        {sukkerd2020tradeoff}
\bibfield{author}{\bibinfo{person}{Roykrong Sukkerd}, \bibinfo{person}{Reid
  Simmons}, {and} \bibinfo{person}{David Garlan}.}
  \bibinfo{year}{2020}\natexlab{}.
\newblock \showarticletitle{Tradeoff-focused contrastive explanation for MDP
  planning}. In \bibinfo{booktitle}{\emph{2020 29th IEEE International
  Conference on Robot and Human Interactive Communication (RO-MAN)}}. IEEE,
  \bibinfo{pages}{1041--1048}.
\newblock


\bibitem[Sutton and Barto(2018)]%
        {sutton2018reinforcement}
\bibfield{author}{\bibinfo{person}{Richard~S Sutton} {and}
  \bibinfo{person}{Andrew~G Barto}.} \bibinfo{year}{2018}\natexlab{}.
\newblock \bibinfo{booktitle}{\emph{Reinforcement learning: An introduction}}.
\newblock \bibinfo{publisher}{MIT press}.
\newblock


\bibitem[Szegedy et~al\mbox{.}(2013)]%
        {szegedy2013intriguing}
\bibfield{author}{\bibinfo{person}{Christian Szegedy},
  \bibinfo{person}{Wojciech Zaremba}, \bibinfo{person}{Ilya Sutskever},
  \bibinfo{person}{Joan Bruna}, \bibinfo{person}{Dumitru Erhan},
  \bibinfo{person}{Ian Goodfellow}, {and} \bibinfo{person}{Rob Fergus}.}
  \bibinfo{year}{2013}\natexlab{}.
\newblock \showarticletitle{Intriguing properties of neural networks}.
\newblock \bibinfo{journal}{\emph{arXiv preprint arXiv:1312.6199}}
  (\bibinfo{year}{2013}).
\newblock


\bibitem[Topin and Veloso(2019)]%
        {topin2019generation}
\bibfield{author}{\bibinfo{person}{Nicholay Topin} {and}
  \bibinfo{person}{Manuela Veloso}.} \bibinfo{year}{2019}\natexlab{}.
\newblock \showarticletitle{Generation of policy-level explanations for
  reinforcement learning}. In \bibinfo{booktitle}{\emph{Proceedings of the AAAI
  Conference on Artificial Intelligence}}, Vol.~\bibinfo{volume}{33}.
  \bibinfo{pages}{2514--2521}.
\newblock


\bibitem[Ustun et~al\mbox{.}(2019)]%
        {ustun2019actionable}
\bibfield{author}{\bibinfo{person}{Berk Ustun}, \bibinfo{person}{Alexander
  Spangher}, {and} \bibinfo{person}{Yang Liu}.}
  \bibinfo{year}{2019}\natexlab{}.
\newblock \showarticletitle{Actionable recourse in linear classification}. In
  \bibinfo{booktitle}{\emph{Proceedings of the conference on fairness,
  accountability, and transparency}}. \bibinfo{pages}{10--19}.
\newblock


\bibitem[van~der Waa et~al\mbox{.}(2018)]%
        {van2018contrastive-exp}
\bibfield{author}{\bibinfo{person}{Jasper van~der Waa},
  \bibinfo{person}{Jurriaan van Diggelen}, \bibinfo{person}{Karel van~den
  Bosch}, {and} \bibinfo{person}{Mark Neerincx}.}
  \bibinfo{year}{2018}\natexlab{}.
\newblock \showarticletitle{Contrastive explanations for reinforcement learning
  in terms of expected consequences}.
\newblock \bibinfo{journal}{\emph{arXiv preprint arXiv:1807.08706}}
  (\bibinfo{year}{2018}).
\newblock


\bibitem[Verma(2019)]%
        {verma2019verifiable}
\bibfield{author}{\bibinfo{person}{Abhinav Verma}.}
  \bibinfo{year}{2019}\natexlab{}.
\newblock \showarticletitle{Verifiable and interpretable reinforcement learning
  through program synthesis}. In \bibinfo{booktitle}{\emph{Proceedings of the
  AAAI Conference on Artificial Intelligence}}, Vol.~\bibinfo{volume}{33}.
  \bibinfo{pages}{9902--9903}.
\newblock


\bibitem[Verma et~al\mbox{.}(2018)]%
        {verma2018programmatically}
\bibfield{author}{\bibinfo{person}{Abhinav Verma},
  \bibinfo{person}{Vijayaraghavan Murali}, \bibinfo{person}{Rishabh Singh},
  \bibinfo{person}{Pushmeet Kohli}, {and} \bibinfo{person}{Swarat Chaudhuri}.}
  \bibinfo{year}{2018}\natexlab{}.
\newblock \showarticletitle{Programmatically interpretable reinforcement
  learning}. In \bibinfo{booktitle}{\emph{International Conference on Machine
  Learning}}. PMLR, \bibinfo{pages}{5045--5054}.
\newblock


\bibitem[Verma et~al\mbox{.}(2020)]%
        {verma2020counterfactual}
\bibfield{author}{\bibinfo{person}{Sahil Verma}, \bibinfo{person}{John
  Dickerson}, {and} \bibinfo{person}{Keegan Hines}.}
  \bibinfo{year}{2020}\natexlab{}.
\newblock \showarticletitle{Counterfactual explanations for machine learning: A
  review}.
\newblock \bibinfo{journal}{\emph{arXiv preprint arXiv:2010.10596}}
  (\bibinfo{year}{2020}).
\newblock


\bibitem[Verma et~al\mbox{.}(2021)]%
        {verma2021counterfactual}
\bibfield{author}{\bibinfo{person}{Sahil Verma}, \bibinfo{person}{John
  Dickerson}, {and} \bibinfo{person}{Keegan Hines}.}
  \bibinfo{year}{2021}\natexlab{}.
\newblock \showarticletitle{Counterfactual Explanations for Machine Learning:
  Challenges Revisited}.
\newblock \bibinfo{journal}{\emph{arXiv preprint arXiv:2106.07756}}
  (\bibinfo{year}{2021}).
\newblock


\bibitem[Vinyals et~al\mbox{.}(2017)]%
        {vinyals2017starcraft}
\bibfield{author}{\bibinfo{person}{Oriol Vinyals}, \bibinfo{person}{Timo
  Ewalds}, \bibinfo{person}{Sergey Bartunov}, \bibinfo{person}{Petko Georgiev},
  \bibinfo{person}{Alexander~Sasha Vezhnevets}, \bibinfo{person}{Michelle Yeo},
  \bibinfo{person}{Alireza Makhzani}, \bibinfo{person}{Heinrich K{\"u}ttler},
  \bibinfo{person}{John Agapiou}, \bibinfo{person}{Julian Schrittwieser},
  {et~al\mbox{.}}} \bibinfo{year}{2017}\natexlab{}.
\newblock \showarticletitle{Starcraft ii: A new challenge for reinforcement
  learning}.
\newblock \bibinfo{journal}{\emph{arXiv preprint arXiv:1708.04782}}
  (\bibinfo{year}{2017}).
\newblock


\bibitem[Vouros(2022)]%
        {vouros2022explainable}
\bibfield{author}{\bibinfo{person}{George~A Vouros}.}
  \bibinfo{year}{2022}\natexlab{}.
\newblock \showarticletitle{Explainable deep reinforcement learning: state of
  the art and challenges}.
\newblock \bibinfo{journal}{\emph{Comput. Surveys}} \bibinfo{volume}{55},
  \bibinfo{number}{5} (\bibinfo{year}{2022}), \bibinfo{pages}{1--39}.
\newblock


\bibitem[Wachter et~al\mbox{.}(2017)]%
        {wachter2017counterfactual}
\bibfield{author}{\bibinfo{person}{Sandra Wachter}, \bibinfo{person}{Brent
  Mittelstadt}, {and} \bibinfo{person}{Chris Russell}.}
  \bibinfo{year}{2017}\natexlab{}.
\newblock \showarticletitle{Counterfactual explanations without opening the
  black box: Automated decisions and the GDPR}.
\newblock \bibinfo{journal}{\emph{Harv. JL \& Tech.}}  \bibinfo{volume}{31}
  (\bibinfo{year}{2017}), \bibinfo{pages}{841}.
\newblock


\bibitem[Wang et~al\mbox{.}(2016)]%
        {wang2016dueling}
\bibfield{author}{\bibinfo{person}{Ziyu Wang}, \bibinfo{person}{Tom Schaul},
  \bibinfo{person}{Matteo Hessel}, \bibinfo{person}{Hado Hasselt},
  \bibinfo{person}{Marc Lanctot}, {and} \bibinfo{person}{Nando Freitas}.}
  \bibinfo{year}{2016}\natexlab{}.
\newblock \showarticletitle{Dueling network architectures for deep
  reinforcement learning}. In \bibinfo{booktitle}{\emph{International
  conference on machine learning}}. PMLR, \bibinfo{pages}{1995--2003}.
\newblock


\bibitem[Wells and Bednarz(2021)]%
        {wells2021explainable}
\bibfield{author}{\bibinfo{person}{Lindsay Wells} {and} \bibinfo{person}{Tomasz
  Bednarz}.} \bibinfo{year}{2021}\natexlab{}.
\newblock \showarticletitle{Explainable ai and reinforcement learning—a
  systematic review of current approaches and trends}.
\newblock \bibinfo{journal}{\emph{Frontiers in artificial intelligence}}
  \bibinfo{volume}{4} (\bibinfo{year}{2021}), \bibinfo{pages}{550030}.
\newblock


\bibitem[Yue et~al\mbox{.}(2018)]%
        {yue2018machine}
\bibfield{author}{\bibinfo{person}{Wenbin Yue}, \bibinfo{person}{Zidong Wang},
  \bibinfo{person}{Hongwei Chen}, \bibinfo{person}{Annette Payne}, {and}
  \bibinfo{person}{Xiaohui Liu}.} \bibinfo{year}{2018}\natexlab{}.
\newblock \showarticletitle{Machine learning with applications in breast cancer
  diagnosis and prognosis}.
\newblock \bibinfo{journal}{\emph{Designs}} \bibinfo{volume}{2},
  \bibinfo{number}{2} (\bibinfo{year}{2018}), \bibinfo{pages}{13}.
\newblock


\bibitem[Zhang et~al\mbox{.}(2020)]%
        {zhang2020explainable}
\bibfield{author}{\bibinfo{person}{Ke Zhang}, \bibinfo{person}{Peidong Xu},
  {and} \bibinfo{person}{Jun Zhang}.} \bibinfo{year}{2020}\natexlab{}.
\newblock \showarticletitle{Explainable AI in deep reinforcement learning
  models: A shap method applied in power system emergency control}. In
  \bibinfo{booktitle}{\emph{2020 IEEE 4th Conference on Energy Internet and
  Energy System Integration (EI2)}}. IEEE, \bibinfo{pages}{711--716}.
\newblock


\bibitem[Zhao et~al\mbox{.}(2009)]%
        {zhao2009reinforcement}
\bibfield{author}{\bibinfo{person}{Yufan Zhao}, \bibinfo{person}{Michael~R
  Kosorok}, {and} \bibinfo{person}{Donglin Zeng}.}
  \bibinfo{year}{2009}\natexlab{}.
\newblock \showarticletitle{Reinforcement learning design for cancer clinical
  trials}.
\newblock \bibinfo{journal}{\emph{Statistics in medicine}}
  \bibinfo{volume}{28}, \bibinfo{number}{26} (\bibinfo{year}{2009}),
  \bibinfo{pages}{3294--3315}.
\newblock


\bibitem[Zrihem et~al\mbox{.}(2016)]%
        {zrihem2016visualizing}
\bibfield{author}{\bibinfo{person}{Nir~Ben Zrihem}, \bibinfo{person}{Tom
  Zahavy}, {and} \bibinfo{person}{Shie Mannor}.}
  \bibinfo{year}{2016}\natexlab{}.
\newblock \showarticletitle{Visualizing dynamics: from t-sne to semi-mdps}.
\newblock \bibinfo{journal}{\emph{arXiv preprint arXiv:1606.07112}}
  (\bibinfo{year}{2016}).
\newblock


\end{thebibliography}


\end{document}